%% file: main.tex
\documentclass{article}

\PassOptionsToPackage{numbers, sort}{natbib}

\usepackage[final]{neurips_2024}

\usepackage[utf8]{inputenc} 
\usepackage[T1]{fontenc}    
\usepackage{hyperref}       
\usepackage{url}            
\usepackage{booktabs}       
\usepackage{amsfonts}       
\usepackage{nicefrac}       
\usepackage{microtype}      
\usepackage{xcolor}         

\usepackage{colortbl}
\usepackage{xcolor}
\definecolor{lightgray}{gray}{0.9}
\usepackage{booktabs}
\usepackage{color, soul}
\usepackage{graphicx}
\usepackage{amsmath}
\usepackage{empheq}

\usepackage{caption}
\usepackage{subcaption} 

\usepackage{algorithm}
\usepackage{algpseudocode}

\usepackage{listings}

\usepackage{pythonhighlight}

\newcommand\freefootnote[1]{%
  \let\svthefootnote\thefootnote
  \let\thefootnote\relax
  \footnotetext{\textsuperscript{*}#1}
  \let\thefootnote\svthefootnote
}

\definecolor{lightgray}{gray}{0.9}
\definecolor{lightblue}{rgb}{0.93,0.95,1.0}
\definecolor{darkgreen}{rgb}{0.0,0.6,0.0}
\definecolor{darkblue}{rgb}{0.0,0.0,0.5}
\definecolor{pinegreen}{rgb}{0.0, 0.47, 0.44}
\definecolor{deepmagenta}{rgb}{0.8, 0.0, 0.8}
\definecolor{amber}{rgb}{1.0, 0.49, 0.0}
\definecolor{Gray}{gray}{0.9}

\newcommand{\ignorebig}[1]{}

\def\Secref#1{Section~\ref{#1}}

\newcommand{\minisection}[1]{\noindent{\textbf{#1}.}}
\newcommand{\tabref}[1]{Table~\ref{#1}}

\newcommand{\figgref}[1]{Figure~\ref{#1}}

\newlength\savewidth

\newcommand{\model}{Multimodal Task Vectors}
\newcommand{\smodel}{MTV}

\definecolor{citecolor}{RGB}{34,139,34}
\definecolor{lightred}{RGB}{241,140,142}
\definecolor{amber(sae/ece)}{rgb}{1.0, 0.49, 0.0}
\definecolor{battleshipgrey}{rgb}{0.52, 0.52, 0.51}
\definecolor{cadmiumorange}{rgb}{0.93, 0.53, 0.18}
\definecolor{applegreen}{rgb}{0.55, 0.71, 0.0}
\definecolor{cadmiumgreen}{rgb}{0.0, 0.42, 0.24}
\definecolor{forestgreen}{rgb}{0.13, 0.55, 0.13}
\definecolor{red}{rgb}{0.89, 0.0, 0.13}

\title{Multimodal Task Vectors Enable Many-Shot Multimodal In-Context Learning}

\author{Brandon Huang\textsuperscript{1*}  \And
 Chancharik Mitra\textsuperscript{1*} \And
 Assaf Arbelle\textsuperscript{2} \And
Leonid Karlinsky\textsuperscript{3} \And
Trevor Darrell\textsuperscript{1} \quad
 Roei Herzig\textsuperscript{1, 2} \\ \\
\textsuperscript{1} University of California, Berkeley \quad \textsuperscript{2} IBM Research \quad \textsuperscript{3} MIT-IBM Watson AI Lab
}

\begin{document}

\maketitle

\input{sec/0_abstract}
\input{sec/1_intro}
\input{sec/2_related_works}

\input{sec/3_methods_v4}
\input{sec/4_eval_and_results}
\input{sec/5_conclusion}



{
    \small
    \bibliographystyle{ieeenat_fullname}

    \bibliography{main}
}

\appendix
\input{sec/7_supplementary}


   

\end{document}

%% file: sec/0_abstract.tex
\begin{abstract}





The recent success of interleaved Large Multimodal Models (LMMs) in few-shot learning suggests that in-context learning (ICL) with many examples can be promising for learning new tasks. However, this \textit{many-shot} multimodal ICL setting has one crucial problem: it is fundamentally limited by the model's context length set at pretraining. The problem is especially prominent in the multimodal domain, which processes both text and images, requiring additional tokens. This motivates the need for a multimodal method to compress many shots into fewer tokens without finetuning. In this work, we enable LMMs to perform multimodal, many-shot in-context learning by leveraging Multimodal Task Vectors (MTV)---compact implicit representations of in-context examples compressed in the model's attention heads. Specifically, we first demonstrate the existence of such MTV in LMMs and then leverage these extracted MTV to enable many-shot in-context learning for various vision-and-language tasks. Our experiments suggest that MTV can scale in performance with the number of compressed shots and generalize to similar out-of-domain tasks without additional context length for inference. Code: \url{https://github.com/Brandon3964/MultiModal-Task-Vector}
\freefootnote{Denotes Equal Contribution}


\end{abstract}



%% file: sec/1_intro.tex
\section{Introduction}

Large Multimodal Models (LMMs) such as GPT-4V~\cite{OpenAI2023GPT4TR}, LLaVA~\cite{liu2023llava, liu2023llava15}, and the BLIP~\cite{instructblip,li2023blip2} family of models demonstrate state-of-the-art performance on a variety of vision and language (VL) tasks due to their strong reasoning capabilities over both text and images. Recent works show that LMMs pre-trained on interleaved text-image data can do multimodal in-context learning~\cite{Bai2023QwenVLAF, Laurenccon2024Idefics2}. In particular, few-shot, in-context learning (ICL) in text-only LLMs has been scaled with an increasing number of examples in long-context language models---a setting called many-shot learning~\cite{Agarwal2024ManyShotIL}. A natural question arises on how to perform many-shot learning in the multimodal domain. 

The first issue with directly applying a many-shot learning regimen to LMMs is the intrinsic limitation of context length. This is especially true in the multimodal domain, as LMMs must encode both text and images, whose embeddings are token-expensive. Moreover, long-context language models, which LMMs leverage for reasoning, struggle to use their entire context length effectively for ICL~\cite{Liu2023LostIT, Li2024LongcontextLS}. Secondly, perhaps due to the misalignment of pretraining tasks with ICL, many instruction-tuned LMMs underperform on tasks in the ICL setting~\cite{Doveh2024TowardsMM-ICL}, suggesting the importance of interleaved LMMs. Finally, there is also the challenge of the increasing memory and run-time required for processing long contexts for every inference call. These challenges motivate a method for compressing multimodal in-context examples into compact, implicit representations. Therefore, in this paper, we propose {\model} ({\smodel})---compact representations of multimodal in-context tasks---within the attention heads of LMMs to enable many-shot ICL. In particular, we show the existence of {\smodel} in interleaved LMMs, and we use them to compress large numbers of multimodal ICL examples.

Recent research in explainability has demonstrated the existence of task vectors in both the language~\cite{Todd2023FunctionVI, Hendel2023InContextLC} and vision~\cite{Hojel2024FindingVT} domains. These task vectors are implicit representations of in-context tasks represented by sets of activations in the model. These activations compactly summarize the information in ICL examples. In our work, we go beyond proving the existence of these task vectors in the multimodal domain by demonstrating their ability to compress examples for many-shot ICL in LMMs without the need for finetuning. 

Our method can be described in three steps. First, given a set of many-shot multimodal ICL examples, we calculate the mean activations corresponding to the last token across multiple inference iterations. Second, to avoid the context length constraint, we select a set of attention heads in the model to store the mean activations of the ICL examples. However, since the downstream task may be zero-shot or use a different number of ICL examples, we select a set of examples aligned with its form. We then use these examples to find an optimal set of LMM head locations where the many-shot examples will be encoded. We refer to these mean activations and locations as {\smodel}, which implicitly encodes the many-shot multimodal examples for use in the downstream task. Finally, for downstream inference, we replace the mean activations from Step 1 with the attention head locations found in Step 2. Since we input examples to the LMM across different iterations in Step 1, {\model} can implicitly encode more examples than are allowable by the context limit. We find that utilizing many examples for extracting {\smodel} surpasses performance on zero-shot and most standard few-shot ICL settings, suggesting the effectiveness of our method. Another key benefit of our method is that it frees up tokens for the model during downstream inference compared to standard few-shot ICL methods. An overview of our method is shown in~\figgref{fig:overview}.

We summarize our main contributions as follows: (i) We show the existence of {\model}, compact implicit representations of in-context functions in LMMs. (ii) {\smodel} can encode more examples than allowed by an LMM's context length, enabling both runtime and memory-efficient multimodal many-shot in-context learning.
(iii) {\smodel} surpasses zero-shot and few-shot ICL settings on various VL benchmarks without finetuning. 
(iv) {\smodel} can scale to larger numbers of examples and can generalize to similar out-of-domain tasks.

%% file: sec/2_related_works.tex
\input{figs/fig1_overview}

\section{Related Works}

\minisection{Many-Shot In-Context Learning} Few-shot in-context learning (ICL) is a significant area of study in text-only LLMs~\cite{Brown2020OGICL,Wei2022EmergentAO}. A natural question arises about the possibility of using a larger number of shots (e.g., hundreds) to further improve performance or learn more complex tasks. Indeed, some early work in text-only \textit{many-shot, in-context learning} suggests performance on different tasks can scale with a larger number of examples~\cite{Bertsch2024InContextLW, Agarwal2024ManyShotIL, Li2023InContextLW, Li2024LongcontextLS}. 

However, scaling ICL in text-only LLMs is a challenge due to the intrinsic context length. One method to increase context length in these models is to apply positional interpolation methods~\cite{Chen2023PosInter, Peng2023YaRNEC}. However, research on these longer-context models finds that they struggle to use the entire context for ICL~\cite{Liu2023LostIT, Li2024LongcontextLS}. Moreover, as inference on long contexts of inputs is also time and memory-expensive, it is unclear whether simply scaling the context of models is practical for enabling multimodal many-shot ICL in open-source models. There is some early evidence of multimodal many-shot ICL being effective in closed-source models~\cite{jiang2024many}, so the question arises as to how to achieve something similar for open-source models. This has led to work that looks to compress explicit input tokens~\cite{Ge2023IncontextAF, Tan2024LLoCOLL, ChevalierAutoCOmp, Mu2023LearningTC, Snell2022LearningBD, Jiang2023LLMLinguaCP}. But crucially, many of these methods require finetuning and only try to preserve performance. Our work is different in that it is the first to enable \textit{multimodal} models with many-shot ICL capabilities, while also improving on complex VL tasks without finetuning.

\minisection{Task Vectors} Our work builds off of research in text-only and vision-only domains showing that internal representations of these models called task vectors~\cite{Todd2023FunctionVI, Hendel2023InContextLC, Hojel2024FindingVT} (or function vectors) can encapsulate tasks outlined by ICL examples. Our is the first demonstration of {\model} ({\smodel}) in LMMs. Going beyond previous work, however, we show that {\smodel} enable LMMs not only to use many-shot, multimodal ICL examples but also scale with more samples, be used alongside explicit ICL shots, and even generalize to unseen classes or similar tasks. 


\minisection{Model Domain Adaptation Methods} 
As LLM and LMM model architectures have advanced, so have methods to allow these models to generalize beyond their pretraining distributions. Methods like instruction tuning~\cite{Wei2021FinetunedLM, liu2023llava, avraham2022svit, sanh2022multitask} have shown strong zero-shot generalization to some out-of-domain tasks, but forgetting remains an issue. One popular solution to this issue involves Parameter Efficient Fine-tuning (PEFT)~\cite{houlsby19adapters}: finetuning either a set of soft prompt input tokens~\cite{Lester2021ThePO, Li2021PrefixTuningOC}, low-rank model weights~\cite{Hu2021LoRALA, Dettmers2023QLoRAEF, Zhang2023AdaLoRA}, or a separate adapter from the main model~\cite{Zhang2023LLaMAAdapter1, Gao2023LLaMAAdapter2, Hu2023LLMAdaptersAA}.

Prompting methods are a well-explored area for adapting models without finetuning. LLM prompting includes zero-shot methods~\cite{Kojima2022LargeLM, Wang2023PlanandSolvePI, Wan2023BetterZR}, few-shot and ICL methods~\cite{Brown2020OGICL,Min2022RethinkingTR,Dong2022ASO,Ma2023FairnessguidedFP}, expert prompting~\cite{Xu2023ExpertPromptingIL}, and Chain-of-Thought (CoT)~\cite{Wei2022ChainOT, Zhang2022AutomaticCO}, with extensions like self-consistency~\cite{Wang2022SelfConsistencyIC}, Tree-of-Thought (ToT)~\cite{Yao2023TreeOT}, and Graph-of-Thought (GoT)~\cite{Besta2023GraphOT, Yao2023BeyondCE, Lei2023BoostingLR} for more complex structures. Similar multimodal prompting methods exist for LMMs as well~\cite{Wang2022LanguageMW, Zheng2023DDCoTDC, Zhang2023MultimodalCR, Wang2023TSciQTM, MitraCCoT}.

\minisection{Large Multimodal Models (LMMs)} The state-of-the-art performance of LMMs~\cite{liu2023llava, liu2023llava15, Alayrac2022FlamingoAV, li2023blip2,instructblip, zhu2023minigpt4, Ye2023mPLUGOwlME, Ye2023mPLUGOwl2RM, Bai2023QwenVLAF, Gong2023MultiModalGPTAV, Driess2023PaLMEAE} on multimodal tasks stems from combining LLMs' reasoning capabilities ~\cite{alfassy2022feta,Raffel2019ExploringTL, Chowdhery2022PaLMSL,herzig2023incorporating,Tay2022UL2UL,Shang2024TraveLERAM} with the perception abilities of vision models. LMMs' generative reasoning also makes them more applicable to complex tasks than previous contrastive methods~\cite{radford2021clip,blip,li2023blip2}. Such tasks include visual question-answering \cite{antol2015vqa, Hudson2019GQAAN, Saikh2022ScienceQAAN, Marino2019OKVQAAV, Jia2021ScalingUV, He2019MomentumCF, Gurari2018VizWizGC} as well as object identification and localization~\cite{Flowers, Birds, Lin2014MSCOCO, krishna2017visual}. Visual Programmatic Models (VPMs) are another class of multimodal methods that makes use of in-context APIs code generation~\cite{Suris2023ViperGPTVI, Ge2023RecursiveVP, Gupta2022VisualPC, Subramanian2023ModularVQ, Schick2023ToolformerLM, Lu2023ChameleonPC, Shen2023HuggingGPTSA, Qin2023ToolLLMFL, Wu2023VisualCT}. However, context length limits both LMMs' and VPMs' ability to use multimodal prompting methods such as ICL~\cite{Brown2020OGICL}. Another key challenge is that many LMMs are pre-trained on single text-image pair data. Recently, many LMM models now pretrain on interleaved text-image data~\cite{Bai2023QwenVLAF, Sun2023Emu2, Alayrac2022FlamingoAV, Laurenccon2024Idefics2, Jiang2024MANTISIM, Doveh2024TowardsMM-ICL, Zhao2023MMICLEV}, making effective multimodal ICL possible. In our work, {\smodel} goes beyond simple few-shot multimodal ICL and scales to many-shot multimodal ICL.


%% file: figs/fig1_overview.tex
\begin{figure}[t!]
    \centering
    \includegraphics[width=.95\linewidth]{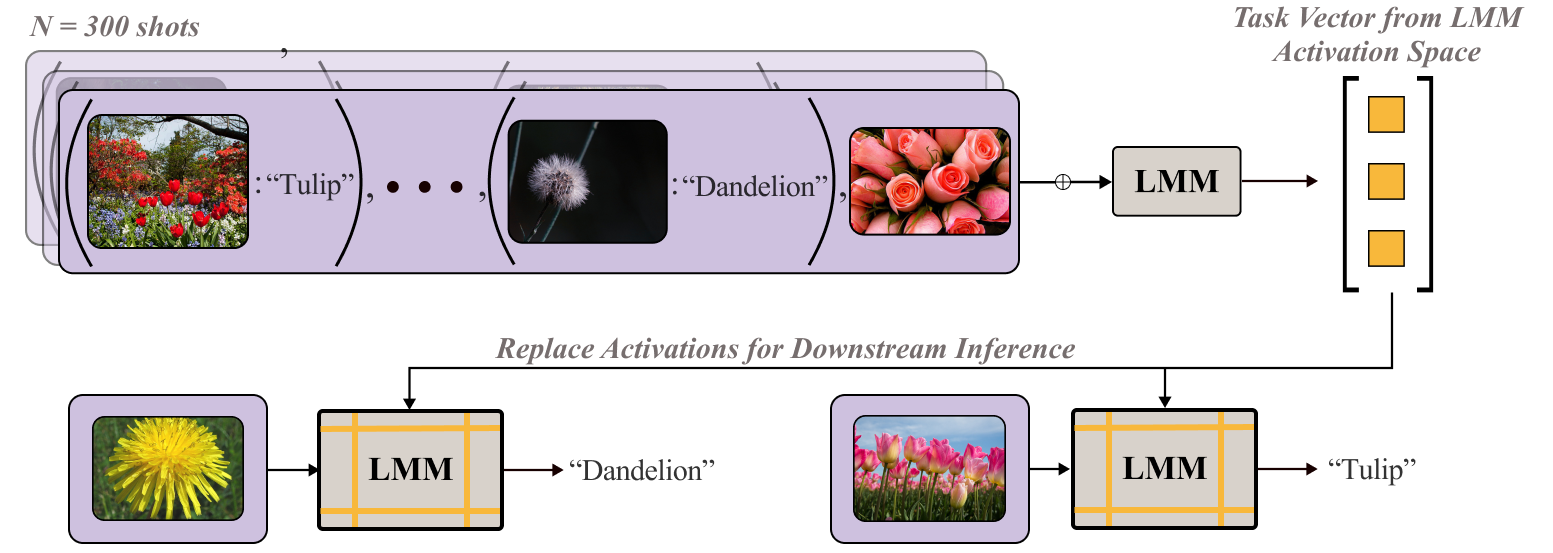}
    \caption{
    \textbf{{\model} ({\smodel}) Overview.}  We overcome an LMM's context length limitation by encoding many shots of multimodal examples as activations in the LMM's latent space. We then directly replace this encoding into the LMM's activation space during downstream inference.
    }
    \label{fig:overview}
\end{figure}


%% file: sec/3_methods_v4.tex
\vspace{-.3cm}
\section{{\model}}
\vspace{-.2cm}
\input{figs/fig1b_detailed}

To address the challenge of performing many-shot multimodal in-context learning, we demonstrate the existence of {\smodel} in LMMs and then leverage them for many-shot multimodal ICL. We begin by describing some background on multimodal ICL and task vectors (\Secref{sec:MTV:preliminaries}). We then introduce our three-step approach: (i) We calculate the mean activations of the attention heads from the many-shot multimodal ICL examples (\Secref{sec:MTV:mu}); (ii) We then extract the set of LMM attention heads locations that best align to the downstream task using an adapted version of the REINFORCE~\cite{Williams2004SimpleSG} algorithm (\Secref{sec:MTV:mtvextract}); and (iii) We replace the calculated mean activation values into the LMM for a downstream task (\Secref{sec:MTV:mtvapp}). The detailed method visual is shown in~\figgref{fig:overview}.

\subsection{Preliminaries}
\label{sec:MTV:preliminaries}
In the multimodal in-context learning setting, an LMM learns a new task outlined by a set of multimodal examples. The input to the LMM would be outlined as follows:
\nolinebreak
\begin{equation}
    I_\mathrm{few} = [(x_1: y_1), (x_2: y_2), \ldots, (x_n: y_n), Q]
\end{equation}
\quad where the model is prompted to answer a query $Q$ given a set of input-output examples (each $x_i$ being a multimodal input and each $y_i$ a text output).

We note that in-context examples are commonly passed sequentially to the LMM, necessarily restricting multimodal ICL to being small numbers of shots due to limited context length. Furthermore, the images require more tokens to embed, which means enabling many-shot ICL is even more challenging in the multimodal domain. To solve this, we utilize our method {\smodel}---which are implicit representations in the model's attention heads that encode a many-shot multimodal ICL task.

We start with a background on task vectors for some task $j$. Given a model $F$, we denote the set of attention-head locations as $\lambda = \{l \mid \forall l \in F \}$ where each location $l$ is indexed as $l = (h, m)$ for the $h^\mathrm{th}$ layer and $m^\mathrm{th}$ attention head. 
Now, task vectors utilize the intermediate outputs of an LMM, called \textbf{activations}. For a given input sequence of written in terms of its tokens $x = \{x_1, x_2, \ldots, x_T\}$, each attention head $(h,m)$ produces an activation $z_l \in \mathbb{R}^{\frac{d}{H}}$ for each token $x_i$, where $d$ is the model's embedding dimension and $H$ is the number of heads. These activations are simply the output vectors of each attention head \textit{before} any linear projection. While each head's activation is typically concatenated with others and projected to form the layer's output, task vectors specifically utilize the pre-projection activations of the final token $x_T$ from each attention head.
We thus define the task vectors as follows: (1) the task vector \textbf{values} $\mu_j$ are a subset of mean activations produced by the attention heads of $F$ given examples of a task, and (2) the task vector \textbf{locations} $\lambda_j$, which denotes a subset of the attention head indices per task. Thus, the task vector is $(\mu_j,\lambda_j)$. For inference, $\mu_j$ replaces the activation values of the heads in the locations given by $\lambda_j$.

In prior work~\cite{Hendel2023InContextLC,Hojel2024FindingVT,Todd2023FunctionVI}, the calculation of the mean activations $\mu_j$ and the extraction of the attention-head locations $\lambda_j$ are used together to extract the task vector. Interestingly, we find that these two steps should be decoupled in order to better align with the downstream task. In our work, we calculate the mean activations $\mu_j$ corresponding to the last token specifically to encode a dataset of many-shot multimodal ICL examples by averaging them across multiple inference calls. However, the downstream task may not always be in the same ICL format as the many-shot examples (e.g., the downstream task uses a different number of shots or is zero-shot). To solve this, we use a separate set of examples that are of the exact format of the downstream task to align the extracted attention-head locations $\lambda_j$ with the inference task. This separation of responsibilities, wherein $\mu_j$ captures the essential information from the many-shot examples and $\lambda_j$ identifies the specific attention head locations for the downstream task, optimizes the utilization of the encoded information at relevant locations within the model.

Our approach to finding {\model} ({\smodel}) $(\mu^\mathrm{MTV}_j, \lambda^\mathrm{MTV}_j)$ allows LMMs to actually leverage many-shot multimodal ICL examples for complex vision-language tasks without being limited by context length. We proceed by first describing how to calculate the mean activations.

\subsection{Step 1: Calculate {\smodel} Mean Activations}
\label{sec:MTV:mu}

The ultimate objective of many-shot multimodal ICL is to use a large number of input-output examples when solving a task $j$. However, it is not trivial to get the LMM to see more examples during inference time than its context length allows.

To address this issue, we pass a few-shot input $I_t$ for each inference call $t$ for a total of $T > 1$ inference calls. Each $I_t$ consists of $N$ shots (where $N > 1$) of multimodal in-context examples in the form of randomly-selected input-output response pairs $(x_t: y_t)$, and $Q_t$, which is the query to be answered by the LMM in that iteration.
\nolinebreak
\begin{equation}
   I_t = [(x_1: y_1), (x_2: y_2), \ldots, (x_N: y_N), Q_t]
\end{equation}
Thus, over $T$ LMM inference calls, we have a many-shot multimodal dataset (of $N \times T$ examples):
\nolinebreak
\begin{align}
   I_\mathrm{many} &= [I_1, I_2,\ldots, I_T] 
\end{align}

However, this dataset is still just a disconnected set of few-shot examples. Next, we would like to connect the separate examples into one unified many-shot multimodal ICL representation.

For each inference call, the LMM is given $N$-shot ICL examples. We calculate the mean of the activations corresponding to the last token of the input $z_{l,j}$ for each attention head index $\forall l \in \lambda$ (\Secref{sec:MTV:preliminaries}) across $T$ inference calls, yielding: 

\begin{equation}
    \forall l \in \lambda: \quad \mu_{l, j} = \frac{1}{T} \sum_{t=1}^T \mathbb{E}[z_{l,j} \mid I_t] = \frac{1}{T} \sum_{t=1}^T \mathbb{E}\left[z_{l,j} \mid (x_1: y_1), (x_2: y_2), \ldots, (x_N: y_N), Q_t\right]
\end{equation}

In this step, we have found the mean activations $\mu_{l,j}$, which encode an internal LMM representation of many shots of multimodal ICL examples. In the next subsection, we describe our methodology for selecting the set of attention heads where these mean activations will be used.

\subsection{Step 2: Extract {\smodel} Attention Head Locations}
\label{sec:MTV:mtvextract}

After Step 1, we now have mean activations for the attention heads of the last token in a given many-shot multimodal task. Yet, we still need to find which set of attention heads $\lambda^\mathrm{MTV}_j$ should be chosen to encode our task.

To choose the set of attention heads, we first prepare a separate set of $S$ examples specifically aligned to the format of the downstream task. For instance, if the downstream setting is a 2-way, one-shot classification task, then the $S$ examples should conform to this paradigm. For our explanation, let's consider a downstream task that is zero-shot such that there is a single query $Q_s$ and corresponding response $R_s$ for all $s \in [1, 2, \ldots, S]$. 

From these examples, we utilize an adapted version of the REINFORCE~\cite{Williams2004SimpleSG} algorithm---an iterative policy optimization method that can be used to find task vector locations~\cite{Hojel2024FindingVT}. Given an LMM $F$, we first select a proposed set of attention head locations by sampling a Bernoulli distribution over the locations multiple times. Next, we directly replace the values of the selected attention heads with the corresponding mean activations $\mu_{l,j}$. Then, after prompting the model with the query $Q_s$, we use the negative cross-entropy loss between the LMM's output logits and the logits of the ground-truth response $R_s$ to optimize the Bernoulli distribution. By optimizing the Bernoulli distribution across $S$ iterations, we are finding the best attention head locations $\lambda^\mathrm{MTV}_j$ for patching in our mean activations. Finally, we can extract $\lambda^\mathrm{MTV}_j$, the optimized indices of attention heads, by sampling our optimized Bernoulli distribution.
\nolinebreak
\begin{equation}
    \lambda^\mathrm{MTV}_j = \mathrm{MTV\_EXTRACT}(F, [Q_1, Q_2, \ldots, Q_S)], [R_1, R_2, \ldots, R_S])
\end{equation}


It is important to note that MTV\_EXTRACT does not require finetuning of the LMM parameters, but rather only inference calls. We describe further the underlying details of our adapted MTV\_EXTRACT algorithm in~\Secref{supp:add_model} of the Supplementary. Having found $\lambda^\mathrm{MTV}_j$ and $\mu_{l,j}$, we describe in what follows, the final procedure to use {\smodel} for inference.

\subsection{Step 3: Multimodal Task Vector Application}
\label{sec:MTV:mtvapp}
After we have identified a set of attention heads $\lambda^\mathrm{MTV}_j$, it is straightforward to apply {\smodel} for inference. We denote the set of mean activations $\mu^\mathrm{MTV}_j$ as follows $\mu^\mathrm{MTV}_j = \{\mu_{l,j} | \forall l \in \lambda^\mathrm{MTV}_j\}$.

To run downstream inference on a new query $Q_\mathrm{new}$ with our model $F$, we directly replace the values of attention heads $\lambda^\mathrm{MTV}_j$ with $\mu^\mathrm{MTV}_j$ and produce the following response $R_\mathrm{new}$:
\nolinebreak
\begin{equation}
    R_\mathrm{new} = F(Q_{new}|\lambda^\mathrm{MTV}_j, \mu^\mathrm{MTV}_j)
\end{equation}
$R_\mathrm{new}$ is thus a response generated using many shots of multimodal examples as implicit context via {\smodel}. The key insight of our method is the importance of $N$ (the number of multimodal examples) and many $T$ (the number of iterations) during the calculation of {\smodel}. This enables an LMM to go beyond its context length to learn more nuanced properties of the task from seeing many examples. Additionally, insertion of {\smodel} directly into the LMM also obviates the need for any context length during downstream inference, actually \textit{freeing} additional context for other use (e.g., an additional prompt, more ICL examples, etc.). Finally, because we align the attention-head locations with the downstream task, {\smodel} can be effectively applied to zero-shot and different ICL settings.



%% file: figs/fig1b_detailed.tex
\begin{figure}[t]
    \centering
    \includegraphics[width=.95\linewidth]{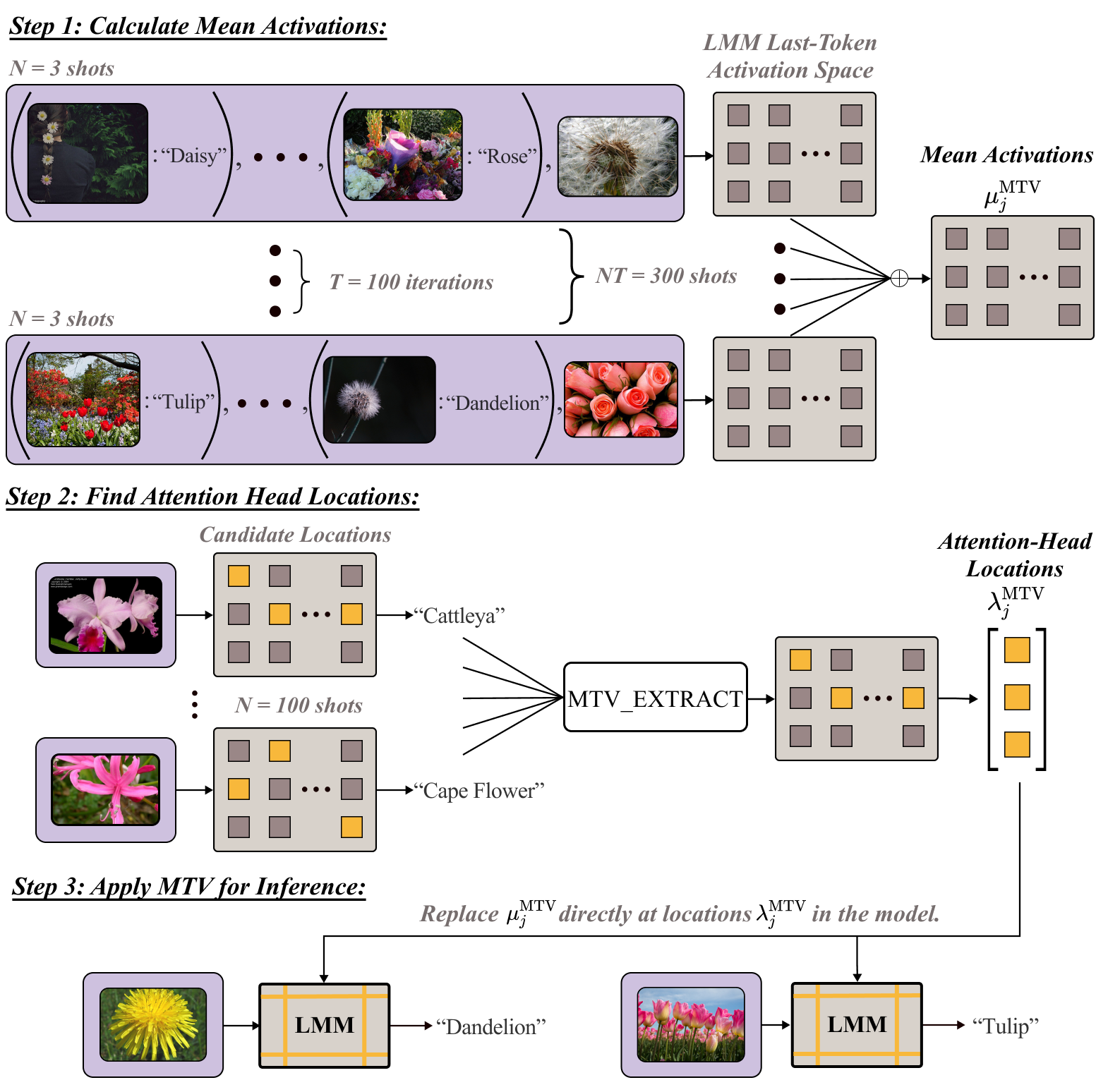}
    \caption{
    \textbf{{\model} ({\smodel}). } In the standard multimodal in-context learning (ICL) paradigm, the number of shots is limited by an LMM's context length. We solve this issue by first finding the mean activations corresponding to the last token of the examples' input (Step 1), and then calculating a set of attention head locations (Step 2) that best align with the downstream task. These mean activations are then replaced directly in these attention head locations (Step 3), enabling many-shot multimodal ICL.
    }
    \label{fig:detailed}
\end{figure}

%% file: sec/4_eval_and_results.tex
\section{Evaluation}
\label{sec:evaluation}
In order for LMMs to perform multimodal ICL, it is important for interleaved data to be included in pretraining. We apply our {\smodel} approach to Qwen-VL~\cite{Bai2023QwenVLAF}, Idefics2-8B~\cite{Laurenccon2023IDEFICS}, and ViLA-1.5-8B~\cite{lin2023vila} three popular interleaved LMMs. For each model, we compare our method to using few-shot ICL  across different vision-and-language tasks like VQA and object identification.

\subsection{Implementation Details}
\label{sec:eval:impl}
We implemented {\smodel} using PyTorch~\cite{paszke2019pytorch}. We used each model's respective official implementation. While the compute and memory requirements differ slightly between models, all our experiments can be run on a single NVIDIA A6000 GPU. For additional information, refer to Supplementary~\Secref{supp:impl}. Our model and weights will be released upon acceptance, and our code is in Supplementary.


\subsection{Models}
In this work, we apply {\smodel} to the following interleaved LMMs as they are better-suited for multimodal ICL as shown by~\cite{Doveh2024TowardsMM-ICL}: (1) \textbf{QwenVL}~\cite{Bai2023QwenVLAF} is a LLaMA-based model that has the ability to process high-resolution images, and its two-stage pre-training methodology, which includes multi-task finetuning and interleaved text-image data. 
(2) \textbf{Idefics2-8B}~\cite{Laurenccon2024Idefics2} is a Mistral-based model that benefits from its pre-training on the expansive OBELICS dataset, which comprises a web-scale collection of interleaved image-text documents. We utilize the base version of the model. This demonstrates multimodal in-context learning abilities. 
(3) \textbf{LLaMA3-ViLA-1.5-8B} (abbreviated as VILA-1.5-8B). ViLA-1.5-8B~\cite{lin2023vila} is an architecture that leverages LLaMA-3 as the LLM backbone. As in others, a significant portion of the model's pretraining data is interleaved text-image data. (4) \textbf{MANTIS-LLaMA3-8B}. MANTIS-LLaMA3-8B~\cite{Jiang2024MANTISIM} is a combination of a SigLIP~\cite{zhai2023sigmoid} visual encoder and LLaMA3~\cite{meta2024llama3} language model finetuned using the MANTIS dataset, a specially curated multi-image dataset that emphasizes co-reference, reasoning, comparing, temporal understanding.

We show the number of tokens per image embedding for each model in~\tabref{tbl:image_embeds} to illustrate the especial importance of MTVs in the image-text domain:

\begin{table}[htbp]
    \centering
    \caption{Per Image Embedding Token Length and Total Context Length for Models}
    \begin{tabular}{l
    >{\columncolor{Gray}}c
    c}
    \toprule
    \textbf{Model Name} & \textbf{\cellcolor{Gray}Per Image Token Length} & \textbf{Total Context Length} \\
    \midrule
    VILA-1.5-8B     & 144  & 8192 \\
    Idefics2-8B & 64   & 8192 \\
    QwenVL   & 256  & 8192 \\
    MANTIS-LLaMA3-8B   & 64  & 8192 \\
    \bottomrule
    \end{tabular}
    \label{tbl:image_embeds}
\end{table}


\input{tbl/tbl1}

\subsection{Datasets}
\label{subsec:data}

We briefly describe the tasks and datasets we evaluate our method on. More details about the datasets and their setup can be found in~\Secref{supp:impl}. 

\minisection{VQA Datasets} We use the following commonly-evaluated datasets which emphasize different aspects of multimodal reasoning, including visual features (VizWiz) and outside knowledge (OK-VQA): (1) \textbf{VizWiz}~\cite{Gurari2018VizWizGC} consists of images taken by visually impaired individuals paired with questions they pose about these images, making it crucial for developing AI systems that assist in real-world, accessibility-focused visual understanding tasks. (2) \textbf{OK-VQA} dataset~\cite{Marino2019OKVQAAV} is designed to push the boundaries of Visual Question Answering (VQA) by focusing on knowledge-based questions, where answers require external knowledge beyond the image content. (3) 

\minisection{Object Classification} We use the following datasets, which are commonly used for object classification in multimodal ICL: (1) The \textbf{Flowers} dataset~\cite{Flowers}, commonly known as the Oxford 102 Flowers dataset, is a collection specifically designed for image-based flower species recognition for fine-grained classification of 102 different categories of flowers. (2) \textbf{Caltech's CUB Dataset on Birds}~\cite{Birds} is a well-known resource for evaluating algorithms on the task of object identification, specifically focused on bird species. It features 200 bird species with roughly 30 images each, annotated with key attributes and bounding boxes. Both Flowers and Birds are formatted as 2-way,1-shot classification episodes, with model inputs being a positive and negative image for the class to be identified in the query image. The response format is a short text response.


\section{Results}
\label{sec:results}
Our main results are shown in~\tabref{tbl:main}. For VQA, we show the results of {\smodel} with 4 shots per 100 iterations to calculate the mean activations and 100 examples for task vector locations (500 examples total). The task vector is extracted using examples from the train set of the dataset and evaluated on the validation set. For object classification, we extract {\smodel} based on a 2-way, one-shot regimen per 100 iterations for both mean activations and task vector locations (200 examples total). The task vector is extracted using a train set of 30\% of the object classes and evaluated on the remaining 70\% of \textit{unseen} classes. 
We demonstrate how {\model} outperforms zero-shot and few-shot ICL settings on three different models on VL tasks, highlighting the effectiveness of our method. 
Next, we describe the unique capabilities of our method, such as scaling to more samples and showing some generalizations to other tasks. More results can be found in~\Secref{supp:more_ablt} of Supplementary.




\subsection{{\smodel} scales with more examples}
We are interested in evaluating (i) the effect of different numbers of shots used \textit{per iteration} to extract MTV and (ii) the effect of different numbers of \textit{iterations} used. We test the impact on accuracy when increasing both of these parameters for QwenVL on the VizWiz validation set. In~\figgref{fig:scaling}, we show on the left that the optimal number of multimodal ICL shots is 16 shots per iteration. Further, we show on the right side of the figure that 1000 examples yield the best performance. These results illustrate that {\smodel} can effectively scale by utilizing larger numbers of ICL examples per iteration and also in aggregate. 
\input{figs/fig2_scaling}
\subsection{{\smodel} works with explicit few-shot examples}
\label{subsec:explicit}
One of the benefits of {\smodel} over a few-shot ICL is the context length that is saved during inference. This is because the many-shot examples are encoded directly in the activation space rather than in the input token space. Thus, we ask whether the LMM can use the freed context for additional few-shot examples. For object classification, we formulate both Flowers and CUB as a 1-shot comparison between a positive and negative sample to identify the correct class (i.e., 2-way, 1-shot ICL by construction). We report results on 1-shot ICL and {\smodel} with 1-shot classification during inference. {\smodel}+1-shot ICL surpasses 1-shot ICL accuracy on these tasks, showing that {\smodel} can be utilized alongside few-shot examples. Furthermore, it is vital to note that the evaluation classes are completely unseen by {\smodel}. Thus, with just a 1-shot ICL example, {\smodel} is able to generalize to unseen classes.

\subsection{{\smodel} heads generalize to other tasks}
\label{subsec:generalize}
In this experiment, we further ask whether the {\smodel} heads $\lambda^{\mathrm{MTV}}_j$ extracted on one task $j$ can generalize to a separate, but similar task $k$. To test this, we use the attention heads extracted from ViLA-1.5-8B on VizWiz for use on OK-VQA. Our results on the left of~\tabref{fig:gen_and_comp} demonstrate that the extracted heads from one task can improve accuracy on another similar task. This generalizability of the heads is significant because it suggests that the heads from {\smodel} may only have to be extracted once to be applied to many other similar tasks. The only calculation necessary then would be the mean activations of the many-shot examples used for the target dataset, making the application of many-shot multimodal ICL even more efficient for similar tasks.

\input{tbl/tbl2}
\subsection{Finetuning as an upper bound}
In~\tabref{subtab:taskvec}, we compare our method to finetuning. To do this, we use finetune on the same number of examples as {\smodel} uses from the train set and evaluate not only on the validation set but also on the validation set of another similar dataset. In particular, for a ViLA-1.5-8B model finetuned on VizWiz, we report accuracy on both VizWiz and OK-VQA validation sets. It can be seen that finetuning is indeed an upper bound on the dataset the model was finetuned on. However, we show that finetuning leads to overfitting on the finetuned dataset and even forgetting the zero-shot capabilities. In contrast, we also show that {\smodel} not only improves zero-shot capabilities but can generalize to similar tasks with only a few inference examples~\tabref{tbl:object_id_specific} and~\tabref{subtab:generalization}.




\subsection{Comparison to other methods} 
We compare our method to two different methods that can find task vectors: Visual Task Vectors (VTV)~\cite{Hojel2024FindingVT} and Function Vectors (FV)~\cite{Todd2023FunctionVI}. Originally, these works could not be applied as-is to support multimodal ICL, but here, we have implemented a version that follows the original exactly with only minor modifications to allow 
performing our evaluated multimodal tasks. More details about the methods can be found in~\Secref{supp:add_model} in the Supplementary. In our experiments~\tabref{subtab:taskvec}, we find that {\smodel} surpasses both methods on VizWiz and OK-VQA. VTV are image-only task vectors that use only one-shot image examples for fixed small $T$ iterations, and they calculate the mean activations and the locations together without aligning to the downstream task. FV are text-only task vectors that use Causal Mediation Analysis~\cite{Pearl2001DirectAI} to extract task vector locations from only the output activations of the last token. The results suggest the importance of finding the task vectors by decoupling the calculation of the mean activations and locations in two separate steps to perform many-shot multimodal ICL more effectively for complex multimodal tasks.

\subsection{Compute and runtime efficiency}
\label{sec:efficiency}
\input{tbl/tbl3}

An important feature of our work is that multimodal ICL examples do not require explicit tokens during inference. Because of this, we are interested in the efficiency gains of our method. Intuitively, the longer {\smodel} extraction time is amortized during downstream inference, where the runtime would be equivalent to the zero-shot case.
Similarly, the memory requirements are maximal during the {\smodel} extraction process but require the same memory as the zero-shot case afterward. In contrast, the ICL tasks have a slower runtime and larger memory requirement throughout due to running inference on $N$ examples \textit{for every iteration}. To demonstrate this, we calculate the maximum memory requirement in gigabytes (GB) for ViLA-1.5-8B on VizWiz using different ICL-shot counts and {\smodel} with 400 examples. As shown in~\tabref{tab:efficiency}, {\smodel} requires less runtime than 16-shot, 8-shot, and 4-shot ICL methods and also requires less memory than 16-shot ICL. These results demonstrate that {\smodel} can encode many multimodal ICL examples with greater efficiency than few-shot methods.





%% file: tbl/tbl1.tex
\definecolor{mediumgray}{gray}{0.8}

\renewcommand{\arraystretch}{1.1}
\newcolumntype{P}[1]{>{\centering\arraybackslash}p{#1}}

\begin{table}[t]
    \caption{\textbf{Results}. (Left) {\smodel} evaluated on VQA datasets. (Right) {\smodel} evaluated on object classification datasets. The baselines are shown in \textcolor{gray}{gray}.}
    \label{tbl:main}
    \begin{subtable}[t]{0.51\textwidth}
         \caption{\textbf{{\smodel} on VQA Benchmarks}}
        \begin{tabular}{m{0.47\textwidth}P{0.14\textwidth}P{0.2\textwidth}}
            \toprule
            \hline
        Model & VizWiz & OK-VQA  \\ \hline
        \rowcolor{gray!30}Flamingo 9B & 28.8 & 44.7 \\ 
        \rowcolor{gray!30}\quad +4-shot ICL& 34.9 & 49.3   \\
        \rowcolor{gray!30}\quad +8-shot ICL &39.4 & 50.0     \\ \hline
        \rowcolor{gray!30}Blip3 & 21.2  & 26.5   \\ 
        \rowcolor{gray!30}\quad +4-shot ICL& 38.4 & 49.2 \\
        \rowcolor{gray!30}\quad +8-shot ICL & 44.3 & 49.1 \\ \hline 
        Qwen-VL-7B & 35.2 &  58.6    \\ 
        \quad +4-shot ICL& 42.0 & \textbf{62.0 }\\
        \quad +8-shot ICL &44.3 &61.5  \\
         \quad +\textbf{{\smodel}} & \textbf{45.6} & \textbf{62.0} \\ \hline
        Idefics2 & 31.3 & 52.4  \\ 
        \quad +4-shot ICL& 40.8 & 51.5   \\
        \quad +8-shot ICL & 43.8& 52.3\\
        \quad +\textbf{{\smodel}} & \textbf{52.5}& \textbf{53.0}\\ \hline
        VILA-1.5-8B & 28.0 & 32.8 \\ 
        \quad +4-shot ICL& 39.3 & 35.6    \\
        \quad +8-shot ICL & 44.2& 36.5 \\
        \quad +\textbf{{\smodel}} & \textbf{55.2}& \textbf{40.6}\\ \hline
        MANTIS-LLaMA3-8B & 36.3 & 51.7 \\ 
        \quad +4-shot ICL& 26.4 & 52.5    \\
        \quad +8-shot ICL & 27.5 & 52.0 \\
        \quad +\textbf{{\smodel}} & \textbf{51.0}& \textbf{52.8}\vspace{0.4mm}\\ \hline
        \bottomrule
        \end{tabular}
       
        \label{tbl:main_results}
    \end{subtable}%
    \begin{subtable}[t]{0.46\textwidth}
        \caption{\textbf{{\smodel} on Object Classification}}
        \centering
        \begin{tabular}{m{0.55\textwidth}P{0.2\textwidth}P{0.12\textwidth}}
        \toprule
        \hline
        Model & Flowers & CUB\\
        \hline
        \rowcolor{gray!30}LLaVA-1.5-13B & &\\
        \rowcolor{gray!30} \quad+ 1-shot ICL & 58.60 & 58.24 \\ \hline
        \rowcolor{gray!30}LLaVA-1.6-13B & &\\
        \rowcolor{gray!30} \quad+ 1-shot ICL& 65.58 & 67.90 \\ \hline
        \rowcolor{gray!30}Flamingo 9B & &\\
        \rowcolor{gray!30} \quad+ 1-shot ICL 9B & 48.78  & 51.2 \\ \hline
        
        \rowcolor{gray!30} IDEFICS-9B & &\\
        \rowcolor{gray!30}\quad+ 1-shot ICL & 55.29  & 62.0 \\  \hline
        \rowcolor{gray!30} Emu 37B & &\\
        \rowcolor{gray!30}\quad+ 1-shot ICL& 52.76 & 53.56 \\  \hline
        Qwen-VL-7B \\ 
        \quad+ 1-shot ICL& 55.0 & 56.5 \\
        \quad+ \textbf{MTV}+1-shot ICL & \textbf{78.1} & \textbf{80.0} \\ \hline
        Idefics2 \\ 
        \quad+ 1-shot ICL& 82.8 & 88.7 \\
        \quad+ \textbf{MTV}+1-shot ICL & \textbf{83.8} & \textbf{89.8} \\ \hline
        VILA-1.5-8B \\ 
        \quad+ 1-shot ICL& 87.4 & 88.4 \\
        \quad+ \textbf{MTV}+1-shot ICL & \textbf{89.3} &\textbf{ 89.7} \\ 
        \hline
        MANTIS-LLaMA3-8B\\ 
        \quad+ 1-shot ICL& 87.4 & 84.0 \\
        \quad+ \textbf{MTV}+1-shot ICL & \textbf{89.8} &\textbf{ 89.7} \\ 
        \hline
        \bottomrule
        \end{tabular}
        
        \label{tbl:object_id_specific}
    \end{subtable}
    
\end{table}

%% file: figs/fig2_scaling.tex
\begin{figure}[htbp]
    \centering
    \begin{minipage}{0.5\textwidth}
        \centering
        \includegraphics[width=\textwidth]{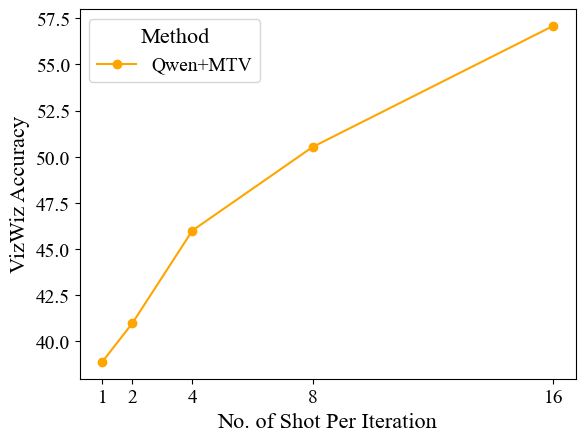} 
    \end{minipage}%
    \begin{minipage}{0.5\textwidth}
        \centering
        \includegraphics[width=\textwidth]{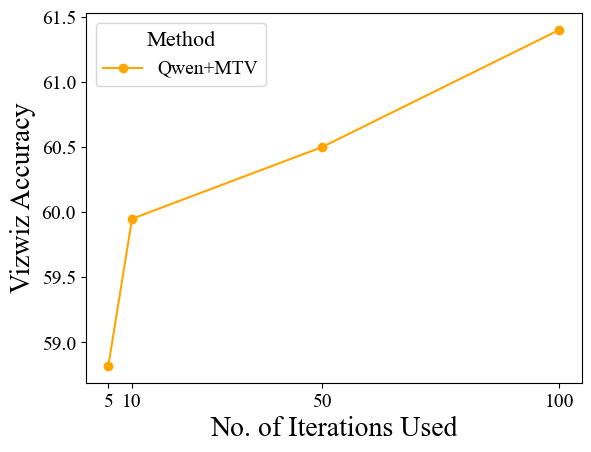} 
    \end{minipage}
    \caption{\textbf{Scaling of Qwen-{\smodel} on VizWiz:} (Left) We show the effect of varying the number of shots per iteration for a fixed 100 iterations. (Right) We also show the effect of varying numbers of iterations fixing 4 shots per iteration.}
    \label{fig:scaling}
\end{figure}

%% file: tbl/tbl2.tex
\begin{table}[htbp]
    \captionsetup{
        justification=raggedright, 
        singlelinecheck=false 
    }
    \caption{\textbf{Generalization \& Method Comparison} (Left) {\smodel}-VizWiz evaluated on OK-VQA. (Right) MTV compared to VizWiz finetuning, function vectors~\cite{Todd2023FunctionVI}, and task vectors~\cite{Hojel2024FindingVT}.}
    \label{fig:gen_and_comp}

     \captionsetup{
    justification=centering, 
    singlelinecheck=false 
    }
    \begin{subtable}[t]{0.49\textwidth}
        \centering
        \subcaption{Attention Head Generalization}
        \resizebox{\textwidth}{!}{%
        \begin{tabular}{@{}l
        >{\columncolor{Gray}}c
        >{\columncolor{Gray}}c
        }
        \toprule
        Model                     & \multicolumn{1}{l}{\cellcolor{Gray}VizWiz}              & \multicolumn{1}{l}{\cellcolor{Gray}OK-VQA}               \\
        \midrule
        ViLA-1.5-8B & 28.0 & 32.8 \\
        \quad+ 4-shot-ICL  \hspace{1cm}  & 39.3& 35.6  \\
        \quad+ 8-shot-ICL   & 44.2 & 36.5  \\
        \quad+ \textbf{{\smodel}-Vizwiz} \hspace{1cm}& \textbf{55.2} & \textbf{38.3} \\
        \bottomrule
        \end{tabular}}
        
        \label{subtab:generalization}
    \end{subtable}%
    \hfill
    \begin{subtable}[t]{0.43\textwidth}
        \centering
        \subcaption{Comparison to Other Methods}
        \resizebox{\textwidth}{!}{%
        \begin{tabular}{@{}l
        >{\columncolor{Gray}}c
        >{\columncolor{Gray}}c
        }
        \toprule
        Model                     & \multicolumn{1}{l}{\cellcolor{Gray}VizWiz}              & \multicolumn{1}{l}{\cellcolor{Gray}OK-VQA} \\
        \midrule
        Qwen-VL-7B & 35.2 & 58.6 \\
        \quad+ VizWiz F.T. \hspace{1cm}  & 62.0 & 25.1 \\
        \quad+ FV   & 36.4 & 59.0 \\
        \quad+ VTV    & 37.0 & 58.9 \\
        \quad+ \textbf{\smodel} & \textbf{45.6} & \textbf{62.0} \\
        \bottomrule
        \end{tabular}}
        
        \label{subtab:taskvec}
    \end{subtable}
\end{table}

\captionsetup{
    justification=raggedright, 
    singlelinecheck=false 
}

%% file: tbl/tbl3.tex
\begin{table}[h]

    \centering

    \begin{tabular}{@{}lcccccc@{}}
    \toprule
    \textbf{Metric} & \textbf{0-shot} & \textbf{4-shot} & \textbf{8-shot} & \textbf{16-shot} & \textbf{MTV (400-shot)} \\ \midrule
    Max GPU Memory (GB) & 17.4 & 18.3 & 19.0 & 20.6 & 19.8 \\
    Runtime per 100 iterations (min) & 1.1 & 2.7 & 3.1 & 3.3 & 1.9 \\
    \bottomrule    \\
    \end{tabular}
    \caption{\textbf{Efficiency:} We show that even though {\smodel} encodes 400 multimodal ICL examples in the mean activations, it still requires less runtime and memory than 8-shot and 16-shot multimodal ICL.
    }
    \label{tab:efficiency}
\end{table}

%% file: sec/5_conclusion.tex
\section{Conclusion}
\label{sec:conclusion}
In this work, we present {\model} a compact, implicit representation that can efficiently encode many-shot multimodal ICL examples for use in complex vision-language tasks. We demonstrate this implicit model representation not only encodes a multimodal ICL task but can also enable many-shot multimodal ICL to surpass zero-shot and few-shot performance on a variety of VL tasks. Our method stands out from previous work in its ability to scale, use additional explicit multimodal ICL examples, and generalize to other similar VL tasks. Our work is a viable way to surpass the limit of context length of an LMM for multimodal ICL and demonstrates clearly that these additional examples aid in multimodal reasoning. Finally, we do not anticipate a specific negative impact, but, as with any Machine Learning method, we recommend exercising caution.

\section{Limitations}
\label{sec:limitations}

While {\model} offers substantial benefits for handling complex vision-language tasks compared to finetuning or few-shot ICL, it is important to recognize certain limitations that accompany our approach. {\smodel} requires access to the internal architecture of an LMM, so while it is an effective solution for all open-source models, its application is restricted from proprietary models, such as GPT-4~\cite{OpenAI2023GPT4TR} and Gemini~\cite{team2023gemini, Reid2024Gemini1.5}. Furthermore, while many-shot ICL is incredibly attractive for many applications, it may not be practical for low-data scenarios where synthetic data~\cite{Agarwal2024ManyShotIL} or the transfer of {\smodel} extracted from another dataset may be required.
We feel these challenges represent great opportunities for future work in the many-shot multimodal in-context learning domain.

\section{Acknowledgements}

We would like to thank Deva Ramanan, Grace Luo, and Suzanne Petryk for their insightful feedback and discussions. This project has received funding from Prof. Darrell’s group in part by DoD, including PTG and/or LwLL programs, as well as BAIR's industrial alliance programs.

%% file: sec/7_supplementary.tex

\clearpage
\setcounter{page}{1}
\maketitlesupplementary

Here, we provide additional information about our experimental results, qualitative examples, implementation details, and datasets. Specifically, \Secref{supp:expr} provides more experiment results,  \Secref{supp:add_model} provides additional method details, \Secref{supp:impl} provides additional implementation details, and \Secref{supp:qual} provides qualitative visualizations to illustrate our approach.


\section{Additional Experiment Results}
\label{supp:expr}

We present several additional experiments that further demonstrate the benefits of our {\smodel} approach. 

\subsection{Additional Experiments}
\label{supp:more_ablt}

Here we provide additional experiments and ablations that further illustrate different characteristics of {\smodel}.

\minisection{Motivation for encoding shots in the activation space.} We highlight our paper’s motivation in addressing the context length token limitation of LMMs by encoding ICL shots in the activation space. An additional limiting factor in the token space is the physical constraints of memory and runtime, which we ablated in~\Secref{sec:efficiency} of the paper. For example, 25-shot ICL is actually the maximum number of vanilla ICL shots that can be run on a single 48GB A6000 GPU for Qwen-VL. We demonstrate the degradation of increasing numbers of multimodal token-space ICL shots (VizWiz-QwenVL) in~\tabref{subtab:icl_degradation}.

\minisection{Effect of shot quality on MTV:}
We assess the connection between textual and activation-space shot quality by comparing MTV using random selection with MTV using high-quality shots selected with the Facility Location algorithm~\cite{schreiber2020apricot}. We apply MTV to QwenVL and use the Qwen GTE embedding model to obtain embeddings for the Facility Location algorithm and present the results in~\tabref{subtab:shot_quality}. Excitingly, we find that high-quality shots do indeed lead to significant improvements in MTV performance.

\minisection{MTV with noisy exemplars.} We compare the robustness of MTV compared to that of vanilla ICL. For QwenVL on VizWiz and OKVQA, we replace 1 of the 4 examples in each iteration of 4-shot ICL and 4-shot-100-iteration MTV with an example from the opposite dataset. We report both accuracy and degradation in~\tabref{subtab:stability}

\begin{table}[htbp]
    \captionsetup{
        justification=raggedright, 
        singlelinecheck=false 
    }
    \caption{\textbf{ICL Degradation, Shot-Quality Impact, and Stability} (Left) Degradation of ICL with increasing number of shots. (Right) Impact of shot quality on MTV and stability of ICL vs MTV with noisy examples.}
    \label{fig:icl_full_comparison}

    \renewcommand{\arraystretch}{1.8}
    \begin{subtable}[t]{0.4\textwidth}
        \centering
        \subcaption{ICL Degradation with Increasing Shots} 
        \begin{tabular}{l c c}
        \toprule
        \textbf{ICL Shots} & \textbf{Acc.} & \textbf{\% Acc. Increase} \\
        \midrule
        0    & 35.2  & -    \\
        4    & 42.0  & 6.8  \\
        8    & 44.3  & 2.3  \\
        16   & 46.9  & 2.6  \\
        20   & 49.0  & 2.1  \\
        25   & 49.8  & 0.8  \\
        \bottomrule
        \end{tabular}
        \label{subtab:icl_degradation}
    \end{subtable}%
    \hfill
    \renewcommand{\arraystretch}{1.24}
    \begin{minipage}[t]{0.5\textwidth}
        \begin{subtable}[t]{\textwidth}
            \centering
            \subcaption{MTV with High-Quality Shots}
            \begin{tabular}{l c}
            \toprule
            \textbf{Model} & \textbf{VizWiz} \\
            \midrule
            QwenVL-7B           & 35.2  \\
            + MTV               & 45.6  \\
            + MTV + F.L. Shots  & 58.1  \\
            \bottomrule
            \end{tabular}
            \label{subtab:shot_quality}
        \end{subtable}%

        \vspace{8pt} 
        \begin{subtable}[t]{\textwidth}
            \centering
            \subcaption{Stability of ICL vs MTV using QwenVL}
            \begin{tabular}{l c c}
            \toprule
            \textbf{} & \textbf{VizWiz} & \textbf{OK-VQA} \\
            \midrule
            4-shot ICL & 41.0 (-1.0) & 61.5 (-0.5) \\
            MTV        & 43.4 (-2.2) & 61.9 (-0.1) \\
            \bottomrule
            \end{tabular}
            \label{subtab:stability}
        \end{subtable}
    \end{minipage}
\end{table}

\minisection{Attention head generalization on object classification tasks~\tabref{subtab:head_gen}} We also test generalization for object classification tasks identical to the formulation described in~\Secref{subsec:generalize}. For clarity, {\smodel} shows another kind of generalization when it is leveraged alongside additional explicit ICL samples. This capability is described in~\Secref{subsec:explicit}. To summarize our experiment, we calculate {\smodel} using the Flowers dataset using 1-shot ICL example for 100 iterations for both the mean activations $\mu^{\mathrm{MTV}}_j$ and the attention head locations $\lambda^{\mathrm{MTV}}_j$. Then, we apply MTV to the CUB task \textit{using the same set of attention head locations from Flowers}. We just calculate the mean activations for the CUB dataset using a 1-shot for 100 iterations (halving our data requirement for this specific scenario). Once again, we find that the heads of {\smodel} can indeed generalize between similar classes.

\input{tbl/tbl4}

\minisection{{\smodel} one-to-one comparison with ICL - ~\tabref{subtab:one_to_one}} Although not directly comparable, we consider an extreme case of {\smodel} where we encode only 4-shots of ICL examples for 1 iteration. This matches the exact setting used in standard 4-shot ICL. Interestingly, {\smodel} applied to both VizWiz and OK-VQA exceeds performance on the 4-shot-ICL case and even {\smodel} formulated on 4-shots per 100 iterations for calculating the mean activations. This result suggests that there may be scope for {\smodel} to be effective in both high and low-data regimens. More research needs to be done to explore this idea.


\input{figs/fig4_flowersscale}

\minisection{Scaling on Flowers Dataset} We provide additional results on the scaling property of {\smodel} on the Flowers dataset. We again note that the examples are \textit{2-way}, one-shot examples with 2 examples (one positive and one negative) for each sample. As in the main paper, we fix 1 shot per iteration to calculate the mean activations, scaling up to 500 total examples used. Our results show that there is a saturation of {\smodel} at 100 examples (i.e., 1 example per 100 iterations). While this still indicates some scaling as the result is an improvement over 20 examples, the results show that the task vector can reach its best accuracy with fewer shots depending on the complexity of the task. Future work to probe more deeply into the scaling nature of {\smodel} across different tasks would be valuable.

\minisection{MTV at extreme shot counts} We delve further into the scaling capabilities by evaluating the performance of MTV at the maximum number of VizWiz shots per iteration allowable by the memory constraints of a single NVIDIA RTX A6000. The experiment shown in \tabref{subtab:extreme} indicates that while MTV does continue to scale, there is also certainly a saturation point for VizWiz. The exact saturation point likely depends on the specific task.

\minisection{Effect of permutation} We consider applying five random seeds to different configurations of {\smodel} comparing its variability under permutation of example order to standard few-shot ICL. We present the mean and standard deviation for these experiments in ~\tabref{subtab:permutation}. Although not significant, in both 4-shot and 8-shot settings, {\smodel} shows less variability to example permutation. This intuitively makes sense as more examples are averaged over multiple iterations, leading to more stable performance across different seeds.
\input{tbl/tbl5}

\minisection{MTV for language-only tasks} While we show the importance of MTV especially for vision-language 
tasks, the methodology can be a powerful way to learn tasks in the language-only domain as well. We demonstrate in~\tabref{tbl:lang_only} the effectiveness of MTV on two common LLM tasks using LLaMA-3-8B~\cite{meta2024llama3}:

\begin{table}[htbp]
    \captionsetup{
        justification=raggedright,
        singlelinecheck=false
    }
    \caption{\textbf{Performance on Language and Document Tasks} (Left) Evaluation on English-Spanish and Antonym Generation tasks. (Right) MTV performance across different shot settings on document tasks.}
    \label{fig:lang_document_tasks}

    \begin{subtable}[t]{0.48\textwidth}
        \centering
        \subcaption{Performance on Language-Only Tasks}
        \resizebox{\textwidth}{!}{%
        \begin{tabular}{l
        >{\columncolor{Gray}}c
        >{\columncolor{Gray}}c}
        \toprule
        & \textbf{English-Spanish} & \textbf{Antonym Generation} \\
        \hline
        10 shot      & 65.2  & 56.0  \\
        400 shot     & 68.5  & 57.6  \\
        \textbf{MTV 4-100}  & \textbf{76.7}  & \textbf{61.7}  \\
        \bottomrule
        \end{tabular}}
        \label{tbl:lang_only}
    \end{subtable}%
    \hfill
    \begin{subtable}[t]{0.48\textwidth}
        \centering
        \subcaption{MTV Document Task Performance}
        \resizebox{\textwidth}{!}{%
        \begin{tabular}{l
        >{\columncolor{Gray}}c
        >{\columncolor{Gray}}c
        >{\columncolor{Gray}}c
        >{\columncolor{Gray}}c}
        \toprule
        & \textbf{0-shot} & \textbf{4-shot} & \textbf{8-shot} & \textbf{MTV} \\
        \hline
        ChartQA  & 19.1  & 25.0  & 26.4  & 34.9  \\
        TextVQA  & 42.4  & 45.4  & 47.1  & 51.0  \\
        \bottomrule
        \end{tabular}}
        \label{tbl:document}
    \end{subtable}
\end{table}


\minisection{MTV on additional document datasets}
Multimodal documents are a form of data with complex compositions of visual and textual modalities, with interleaved language, photograph, and chart information. As such, we provide some preliminary results on the effectiveness of MTV on these types of datasets in~\tabref{tbl:document}. These encouraging results prompt future research into the domain of leveraging task vectors for learning challenging document tasks.

Here we provide some additional method details about {\smodel}, Visual Task Vectors (VTV)~\cite{Hojel2024FindingVT}, and Function Vectors~\cite{Todd2023FunctionVI} (FV).

\subsection{MTV-EXTRACT}
\label{supp:add_model}
We describe the particulars of our MTV-EXTRACT algorithm for finding the set of attention head locations that best align with the downstream task as follows ($Q_s$ and $R_s$ are formatted identically to the downstream task):

\begin{algorithm}
\caption{MTV-EXTRACT for finding task vector locations}
\label{alg:mtv_extract}
\begin{algorithmic}[1]
\Require $F$ (LMM), $S$ (examples), $\mu_j$ (mean activations), $Q_s, R_s$ (queries and responses)
\Ensure $\lambda^\mathrm{MTV}_j$ (optimized attention head locations)

\State Initialize $\theta$ randomly
\For{$s \gets 1$ to $S$}
    \For{$i \gets 1$ to 32} \Comment{Sampling heads 32 times}
        \State Sample $\lambda_i \sim \text{Bernoulli}(\sigma(\theta))$
        \State Replace activations for $\lambda_i$ in $F$ with $\mu_{l,j}$
        \State Compute output logits $O_s \gets F(Q_s)$ \Comment{Pass $Q_s$ to LMM $F$}
        \State $L_i \gets$ Negative Cross-Entropy($O_s$, $R_s$)
    \EndFor
    \State $\theta \gets \text{Adam}(\theta, \nabla_{\theta} \frac{1}{32} \sum_{i=1}^{32} L_i)$ \Comment{Update rule}
\EndFor
\State Sample final $\lambda^\mathrm{MTV}_j \sim \text{Bernoulli}(\sigma(\theta))$ \Comment{Final set of head locations}
\State \Return $\lambda^\mathrm{MTV}_j$
\end{algorithmic}
\end{algorithm}

We point out a few important factors. It is important to note that none of the parameters of $F$ are being finetuned through any gradient update. We take the negative cross-entropy (negative as MTV\_EXTRACT draws inspiration from REINFORCE~\cite{Williams2004SimpleSG}, which is a policy optimization algorithm) between the output logits $O_s$ and the first token of the target response $R_s$ for a simple update scheme. This along with the choice of 32 samples of the Bernoulli distribution are ones we encourage more experimentation with in future work. 
\subsection{Visual TaskVectors (VTV) Adaptation for Multimodal ICL}

Visual Task Vectors (VTV)~\cite{Hojel2024FindingVT} were originally designed to be applied to large vision-transformer-based models. We make as few changes as possible to apply this method for multimodal tasks. We preserve VTVs distinct factors like a the usage of 1-shot examples for both calculation of the mean activations and attention head locations regardless of the format of the downstream task. Furthermore, we fix the number of iterations for both mean activation and attention head calculation at 10. Finally, we replace the proposed MSE loss with a cross-entropy loss that is more suited for an LMM task. 

\subsection{Function Vectors (FV)}
Because Function Vectors describe text-only task vectors, we follow the implementation of Function Vectors~\cite{Todd2023FunctionVI} almost exactly as LLMs and LMMs are similar. The only major change made is the use of many-shot multimodal ICL examples for mean activation calculation. We preserve the lack of an optimization method for the layer used to replace the mean activations. Rather than performing a standard grid search over the set of layers, we set the layer number to 20 as recommended for LLaMA and LLaMA-based models by the paper. The only other difference is the encoding of multimodal ICL examples. Again due the the similarity between LMMs and text-only LLMs, these tests can be used as needed as long as the multimodal inputs are properly processed by the LMM.

\section{Additional Implementation Details}
\label{supp:impl}
To run all of our experiments, we use 1 NVIDIA RTX 6000 GPU. Importantly, this includes the runtime and efficiency ablations, which were evaluated on the same GPU for consistency. Please refer to the respective model's paper for their specific implementation details of the architecture. Besides the output token generation length, which varies depending on the standard setting for each task, we use the default generation parameters (e.g. temperature and no. of beams in beam search) recommended for each model. In the following sections, we describe some of the finer nuances of our MTV-EXTRACT process as well as our implementations of the Visual Task Vectors (VTV) and Function Vectors (FV) implementations.

\subsection{VizWiz}
\minisection{Dataset} The VizWiz dataset is designed to challenge and evaluate the capabilities of Large Multimodal Models (LMMs) in understanding and responding to real-world visual questions. This dataset is comprised of images accompanied by spoken questions, which have been transcribed and paired with answers. Each image in this dataset is sourced from visually impaired individuals seeking assistance, thereby incorporating a wide array of everyday challenges they face. This setup is inherently diverse and often requires high-level visual understanding combined with contextual reasoning, making them a robust benchmark for assessing the practical utility of LMMs in assistive technologies. The format of the dataset samples is an image paired with a text question. The LMM is required to provide a short response limited to 10 tokens or respond with ``unanswerable" if the question is not answerable give the image.

For this research paper, we specifically utilize the VizWiz dataset to benchmark the performance of our proposed task vectors in multimodal in-context learning (MM-ICL) on a dataset that challenges visual scene understanding of LMMs. We extract {\smodel} on the training set and evaluate on the evaluation set containing 4,319 validation image/question pairs. 

\minisection{Inference details} We use the standard VQA question-answer response format that is outlined in the QwenVL repository~\url{https://github.com/QwenLM/Qwen-VL}. Put simply, the LMM is presented with an image and a corresponding text question. The response is then expected in a short text format of no more than 10 tokens (set as the ``max\_tokens'' parameter in the LMM). One nuance is the special answer ``unanswerable". We handle this by providing {\smodel} and all baselines with the following prompt for every question: ``First carefully understand the given examples. Then use the given image and answer the question in the same way as the examples. If the question can not be answered, respond unanswerable. " The official dataset can be downloaded at~\url{https://vizwiz.org/tasks-and-datasets/vqa/}.

\subsection{OK-VQA}

\minisection{Dataset} The OK-VQA dataset, differs from traditional VQA datasets in its focus on necessitating knowledge beyond what is presented in the given images. This dataset encompasses over 14,000 questions that are not merely reliant on visual cues but require associative reasoning with external data sources, making it a unique tool for evaluating AI's capability in handling complex, knowledge-driven queries. Thus, we evaluate on this dataset to test whether {\smodel} can be beneficial for this type of reasoning.

We once again extract {\smodel} on the train set and evaluate on the validation set. OK-VQA is formatted as an image with a corresponding text question. However, it is important to note that the text question heavily relies on external knowledge to answer. Examples of questions can be found in~\Secref{supp:qual}.

\minisection{Inference details} We use the standard VQA question-answer response format that is outlined in the QwenVL repository~\url{https://github.com/QwenLM/Qwen-VL}. Put simply, the LMM is presented with an image and a corresponding text question. The response is then expected in a short text format of no more than 10 tokens (set as the ``max\_tokens'' parameter in the LMM). We do not add any additional prompts or special tokens apart from prompt format or image tokens required by the model being evaluated. The official dataset can be downloaded at~\url{https://okvqa.allenai.org/}.






\subsection{Flowers}
\minisection{Dataset}
Flowers~\cite{Flowers} is an object classification dataset that requires fine-grained classification of 102 different flower species. The Flowers dataset is formulated as a 2-way, 1-shot task where one example is the positive sample and the other is the negative sample. In this way, the data poses a unique challenge for {\smodel} having to store examples with two associated images. Thus, given the 2-way examples and the query image, the LMM is tasked with selecting the correct class from the given two options. Examples can be found in~\Secref{supp:qual}

\minisection{Implementation Details} We use the official data released by the authors which is available at \url{https://www.robots.ox.ac.uk/~vgg/data/flowers/}. We provide a Python code snippet below showing the Flowers data format:
\lstset{
basicstyle=\small\ttfamily,
columns=flexible,
breaklines=true
}
\begin{lstlisting}
def format_flower(cur_data):
    pos = cur_data["pos"]
    neg = cur_data["neg"]
    pos_label = cur_data["pos_label"]
    neg_label = cur_data["neg_label"]
    query = cur_data["query"]
    rand_num = random.randint(0,1)
    if rand_num == 0:
        pos_example = f"<img>{pos}</img>What is the type of flower in the image? A.{pos_label} B.{neg_label}\nAnswer with the option's letter from the given choice directly. Answer: A\n"
        
        neg_example = f"<img>{neg}</img>What is the type of flower in the image? A.{pos_label} B.{neg_label}\nAnswer with the option's letter from the given choice directly. Answer: B\n"
        
        cur_query = f"<img>{query}</img>What is the type of flower in the image? A.{pos_label} B.{neg_label}\nAnswer with the option's letter from the given choice directly. Answer:"
        query_label = "A"
        return pos_example + neg_example + cur_query, query_label, -1
        
    else:
        pos_example = f"<img>{pos}</img>What is the type of flower in the image? A.{neg_label} B.{pos_label}\nAnswer with the option's letter from the given choice directly. Answer: B\n"
        
        neg_example = f"<img>{neg}</img>What is the type of flower in the image? A.{neg_label} B.{pos_label}\nAnswer with the option's letter from the given choice directly. Answer: A\n"
        
        cur_query = f"<img>{query}</img>What is the type of flower in the image? A.{neg_label} B.{pos_label}\nAnswer with the option's letter from the given choice directly. Answer:"
        query_label = "B"
        return neg_example + pos_example + cur_query, query_label, -1
        
\end{lstlisting}

\subsection{CUB}
\minisection{Dataset}
CUB~\cite{Birds} or CUB-200-2011 is an object classification dataset that tests the fine-grained classification of 200 classes of birds. Similar to the Flowers dataset, CUB is formulated as a 2-way, 1-shot task where one example is the positive sample and the other is the negative sample. In this way, the data poses a unique challenge for {\smodel} having to store examples with two associated images. Thus, given the 2-way examples and the query image, the LMM is tasked with selecting the correct class from the given two options. 

\minisection{Implementation Details} We use the official data released by the authors which is available at \url{https://www.vision.caltech.edu/datasets/cub_200_2011/}. We provide a Python code snippet below showing the Flowers data format:

\begin{lstlisting}
def format_cub(cur_data):
    pos = cur_data["pos"]
    neg = cur_data["neg"]
    pos_label = cur_data["pos_label"]
    neg_label = cur_data["neg_label"]
    query = cur_data["query"]
    rand_num = random.randint(0,1)
    if rand_num == 0:
        pos_example = f"<img>{pos}</img>What is the type of bird in the image? A.{pos_label} B.{neg_label}\nAnswer with the option's letter from the given choice directly. Answer: A\n"
        
        neg_example = f"<img>{neg}</img>What is the type of bird in the image? A.{pos_label} B.{neg_label}\nAnswer with the option's letter from the given choice directly. Answer: B\n"
        
        cur_query = f"<img>{query}</img>What is the type of bird in the image? A.{pos_label} B.{neg_label}\nAnswer with the option's letter from the given choice directly. Answer:"
        query_label = "A"
        return pos_example + neg_example + cur_query, query_label, -1
        
    else:
        pos_example = f"<img>{pos}</img>What is the type of bird in the image? A.{neg_label} B.{pos_label}\nAnswer with the option's letter from the given choice directly. Answer: B\n"
        
        neg_example = f"<img>{neg}</img>What is the type of bird in the image? A.{neg_label} B.{pos_label}\nAnswer with the option's letter from the given choice directly. Answer: A\n"
        
        cur_query = f"<img>{query}</img>What is the type of bird in the image? A.{neg_label} B.{pos_label}\nAnswer with the option's letter from the given choice directly. Answer:"
        query_label = "B"
        return neg_example + pos_example + cur_query, query_label, -1
 
\end{lstlisting}

\section{Qualitative Visualizations}

\label{supp:qual}

We present further qualitative success and failure cases of \textbf{QwenVL-{\smodel}} in Figure~\ref{fig:examples} on OK-VQA and Flowers.
\input{figs/fig5_examples}
\section{Licenses and Privacy}
\label{supp:datasets:Licenses}
The license, PII, and consent details of each dataset are in the respective papers. In addition, we wish to emphasize that the datasets we use do not contain any harmful or offensive content, as many other papers in the field also use them. Thus, we do not anticipate a specific negative impact, but, as with any machine learning method, we recommend exercising caution. 

\clearpage

\section*{NeurIPS Paper Checklist}

\begin{enumerate}

\item {\bf Claims}
    \item[] Question: Do the main claims made in the abstract and introduction accurately reflect the paper's contributions and scope?
    \item[] Answer: \answerYes{} 
    \item[] Justification: Yes, the main claims are supported both by the main results and additional ablations and experiments in Sections~\ref{sec:results}

\item {\bf Limitations}
    \item[] Question: Does the paper discuss the limitations of the work performed by the authors?
    \item[] Answer: \answerYes{} 
    \item[] Justification: We discussed about limitations in Section~\ref{sec:limitations}.

\item {\bf Theory Assumptions and Proofs}
    \item[] Question: For each theoretical result, does the paper provide the full set of assumptions and a complete (and correct) proof?
    \item[] Answer: \answerNA{} 
    \item[] Justification: This point is not relevant; it is not a theory paper.

    \item {\bf Experimental Result Reproducibility}
    \item[] Question: Does the paper fully disclose all the information needed to reproduce the main experimental results of the paper to the extent that it affects the main claims and/or conclusions of the paper (regardless of whether the code and data are provided or not)?
    \item[] Answer: \answerYes{} 
    \item[] Justification: Everything is reproducible. We include any necessary details in Section~\ref{sec:evaluation} as well as the Supplemental sections.

\item {\bf Open access to data and code}
    \item[] Question: Does the paper provide open access to the data and code, with sufficient instructions to faithfully reproduce the main experimental results, as described in supplemental material?
    \item[] Answer: \answerYes{} 
    \item[] Justification: Yes, code is provided in the abstract.

\item {\bf Experimental Setting/Details}
    \item[] Question: Does the paper specify all the training and test details (e.g., data splits, hyperparameters, how they were chosen, type of optimizer, etc.) necessary to understand the results?
    \item[] Answer: \answerYes{} 
    \item[] Justification: Yes, all is included in Section~\ref{sec:evaluation} as well as the provided code in the abstract.

\item {\bf Experiment Statistical Significance}
    \item[] Question: Does the paper report error bars suitably and correctly defined or other appropriate information about the statistical significance of the experiments?
    \item[] Answer: \answerNA{} 
    \item[] Justification: Our paper does not require error bars or statistical significance, only accuracy.

\item {\bf Experiments Compute Resources}
    \item[] Question: For each experiment, does the paper provide sufficient information on the computer resources (type of compute workers, memory, time of execution) needed to reproduce the experiments?
    \item[] Answer: \answerYes{} 
    \item[] Justification: Yes, we describe required compute for our method in Section~\ref{sec:evaluation}.
    
\item {\bf Code Of Ethics}
    \item[] Question: Does the research conducted in the paper conform, in every respect, with the NeurIPS Code of Ethics \url{https://neurips.cc/public/EthicsGuidelines}?
    \item[] Answer: \answerYes{} 
    \item[] Justification: We followed the NeurIPS Code of Ethics.

\item {\bf Broader Impacts}
    \item[] Question: Does the paper discuss both potential positive societal impacts and negative societal impacts of the work performed?
    \item[] Answer: \answerYes{} 
    \item[] Justification: Broader impacts are disccused in Section~\ref{sec:conclusion}.
    
\item {\bf Safeguards}
    \item[] Question: Does the paper describe safeguards that have been put in place for responsible release of data or models that have a high risk for misuse (e.g., pretrained language models, image generators, or scraped datasets)?
    \item[] Answer: \answerNo{} 
    \item[] Justification: We don't have any safeguards to discuss here.

\item {\bf Licenses for existing assets}
    \item[] Question: Are the creators or original owners of assets (e.g., code, data, models), used in the paper, properly credited and are the license and terms of use explicitly mentioned and properly respected?
    \item[] Answer: \answerNA{} 
    \item[] Justification: All data and code are credited.

\item {\bf New Assets}
    \item[] Question: Are new assets introduced in the paper well documented and is the documentation provided alongside the assets?
    \item[] Answer: \answerNA{}. 
    \item[] Justification: We present a method and no additional assets are introduced in the paper.

\item {\bf Crowdsourcing and Research with Human Subjects}
    \item[] Question: For crowdsourcing experiments and research with human subjects, does the paper include the full text of instructions given to participants and screenshots, if applicable, as well as details about compensation (if any)? 
    \item[] Answer: \answerNo{} 
    \item[] Justification: No human subjects or experimental data involving humans was used in this research.

\item {\bf Institutional Review Board (IRB) Approvals or Equivalent for Research with Human Subjects}
    \item[] Question: Does the paper describe potential risks incurred by study participants, whether such risks were disclosed to the subjects, and whether Institutional Review Board (IRB) approvals (or an equivalent approval/review based on the requirements of your country or institution) were obtained?
    \item[] Answer: \answerNo{} 
    \item[] Justification: This research does not have any experiments with human subjects.

\end{enumerate}





%% file: tbl/tbl4.tex
\begin{table}[htbp]
    \captionsetup{
        justification=raggedright, 
        singlelinecheck=false 
    }
    \caption{\textbf{Generalization \& Direct ICL Comparison} (Left) {\smodel}-Flowers evaluated on OK-VQA. (Right) Direct comparison of MTV extracted from 4-shots, 1-iteration (MTV\_4shot\_1it) compared to 4-shot ICL}
    \label{fig:gen_2}

     \captionsetup{
    justification=centering, 
    singlelinecheck=false 
    }
    \begin{subtable}[t]{0.49\textwidth}
        \centering
        \subcaption{Attention Head Generalization}
        \resizebox{\textwidth}{!}{%
        \begin{tabular}{@{}l
        >{\columncolor{Gray}}c
        >{\columncolor{Gray}}c
        }
        \toprule
        Model                     & \multicolumn{1}{l}{\cellcolor{Gray}Flowers}              & \multicolumn{1}{l}{\cellcolor{Gray}CUB}               \\
        \midrule
        ViLA-1.5-8B &&   \\
        \quad+ 1-shot-ICL  \hspace{1cm} & 87.4 & 88.4 \\
        \quad+ \textbf{{\smodel}-Flowers}+1-shot-ICL \hspace{1cm}& 89.3 &\textbf{ 89.9} \\
        \bottomrule
        \end{tabular}}
        
        \label{subtab:head_gen}
    \end{subtable}%
    \hfill
    \begin{subtable}[t]{0.43\textwidth}
        \centering
        \subcaption{Comparison to Other Methods}
        \resizebox{\textwidth}{!}{%
        \begin{tabular}{@{}l
        >{\columncolor{Gray}}c
        >{\columncolor{Gray}}c
        }
        \toprule
        Model                     & \multicolumn{1}{l}{\cellcolor{Gray}VizWiz}              & \multicolumn{1}{l}{\cellcolor{Gray}OK-VQA} \\
        \midrule
        ViLA-1.5-8B & 28.0 & 32.8 \\
        \quad+ 4-shot-ICL  \hspace{1cm}  & 39.3& 35.6  \\
        \quad+ \textbf{\smodel}{\_4shot\_1it} & 57.4 & 40.0 \\
        \bottomrule
        \end{tabular}}
        
        \label{subtab:one_to_one}
    \end{subtable}
\end{table}

\captionsetup{
    justification=raggedright, 
    singlelinecheck=false 
}

%% file: figs/fig4_flowersscale.tex
\begin{figure}[h]
    \centering
    \includegraphics[width=0.60\linewidth]{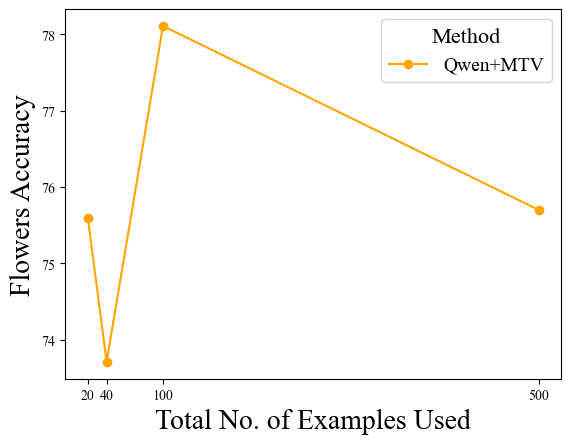}
    \caption{
    \textbf{Efficiency.} We show that for Flowers, {\smodel} does scale to but only up to 100 examples in our experiments.
    }
    \label{fig:flowers}
\end{figure}

%% file: tbl/tbl5.tex



\begin{table}[htbp]
    \captionsetup{
        justification=raggedright,
        singlelinecheck=false
    }
    \caption{\textbf{QwenVL-7B: Generalization \& Variability to Permutation} (Left) Evaluation of MTV on extreme shot-iteration counts. (Right) Variability to permutation across 5 seeds.}
    \label{fig:extreme_permutation}

    \begin{subtable}[t]{0.48\textwidth}
        \centering
        \subcaption{QwenVL-7B: Extreme shot-count performance}
        \resizebox{\textwidth}{!}{%
        \begin{tabular}{l
        >{\columncolor{Gray}}c
        }
        \toprule
        Model                     & \cellcolor{Gray}Accuracy (\%) \\
        \midrule
        QwenVL-7B &  \\
        \quad\quad + MTV\_20shot\_100it& 54.9 \\
        \quad\quad + MTV\_20shot\_200it & 55.1 \\
        \quad\quad + MTV\_25shot\_100it & 56.4 \\
        \quad\quad + MTV\_25shot\_200it & 51.4 \\
        \bottomrule
        \end{tabular}}
        \label{subtab:extreme}
    \end{subtable}%
    \hfill
    \begin{subtable}[t]{0.50\textwidth}
        \centering
        \subcaption{Permutation Variability}
        \resizebox{\textwidth}{!}{%
        \begin{tabular}{l
        >{\columncolor{Gray}}c}
        \toprule
        Model                     & \cellcolor{Gray}Accuracy (\%) \\
        \midrule
        QwenVL-7B &  \\
        \quad\quad + MTV\_4-shot\_100it & 45.2 (0.7) \\
        \quad\quad + MTV\_4-shot\_200it & 48.3 (0.4) \\
        \quad\quad + MTV\_8-shot\_100it & 50.4 (0.9) \\
        \quad\quad + MTV\_8-shot\_200it & 51.8 (0.6) \\
        \quad\quad + 4-shot ICL & 41.3 (0.8) \\
        \quad\quad +  8-shot ICL & 42.9 (1.5) \\
        \bottomrule
        \end{tabular}}
        \label{subtab:permutation}
    \end{subtable}
\end{table}


%% file: figs/fig5_examples.tex
\begin{figure}[h]
    \centering
    \includegraphics[width=\linewidth]{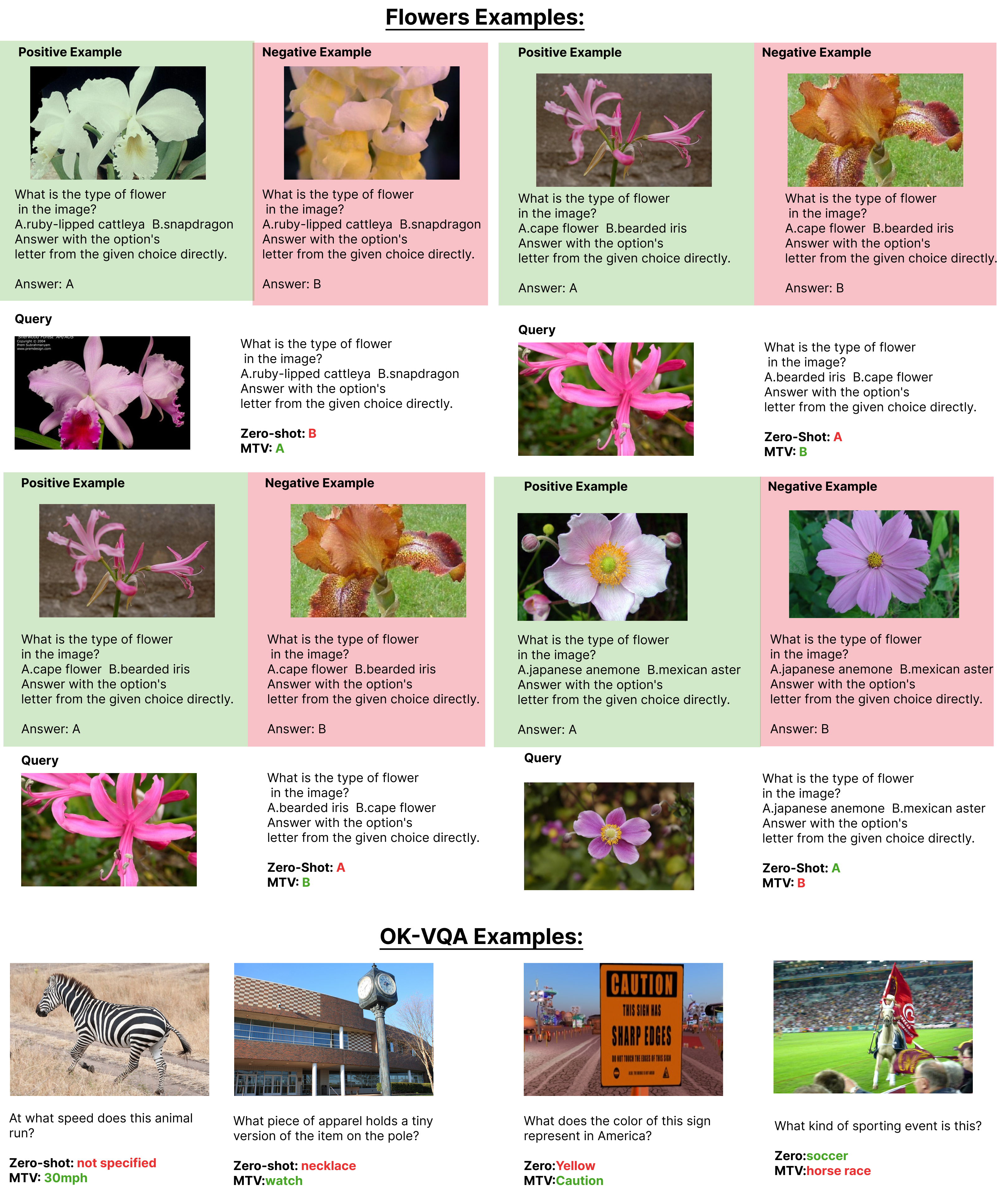}
    \caption{
    \textbf{Efficiency.} We show that for Flowers, {\smodel} does scale to but only up to 100 examples in our experiments.
    }
    \label{fig:examples}
\end{figure}

%% file: main.bbl
\begin{thebibliography}{105}
\providecommand{\natexlab}[1]{#1}
\providecommand{\url}[1]{\texttt{#1}}
\expandafter\ifx\csname urlstyle\endcsname\relax
  \providecommand{\doi}[1]{doi: #1}\else
  \providecommand{\doi}{doi: \begingroup \urlstyle{rm}\Url}\fi

\bibitem[Agarwal et~al.(2024)Agarwal, Singh, Zhang, Bohnet, Chan, Anand, Abbas, Nova, Co-Reyes, Chu, Behbahani, Faust, and Larochelle]{Agarwal2024ManyShotIL}
Rishabh Agarwal, Avi Singh, Lei~M. Zhang, Bernd Bohnet, Stephanie Chan, Ankesh Anand, Zaheer Abbas, Azade Nova, John~D. Co-Reyes, Eric Chu, Feryal M.~P. Behbahani, Aleksandra Faust, and Hugo Larochelle.
\newblock Many-shot in-context learning.
\newblock 2024.

\bibitem[Alayrac et~al.(2022)Alayrac, Donahue, Luc, Miech, Barr, Hasson, Lenc, Mensch, Millican, Reynolds, Ring, Rutherford, Cabi, Han, Gong, Samangooei, Monteiro, Menick, Borgeaud, Brock, Nematzadeh, Sharifzadeh, Binkowski, Barreira, Vinyals, Zisserman, and Simonyan]{Alayrac2022FlamingoAV}
Jean-Baptiste Alayrac, Jeff Donahue, Pauline Luc, Antoine Miech, Iain Barr, Yana Hasson, Karel Lenc, Arthur Mensch, Katie Millican, Malcolm Reynolds, Roman Ring, Eliza Rutherford, Serkan Cabi, Tengda Han, Zhitao Gong, Sina Samangooei, Marianne Monteiro, Jacob Menick, Sebastian Borgeaud, Andy Brock, Aida Nematzadeh, Sahand Sharifzadeh, Mikolaj Binkowski, Ricardo Barreira, Oriol Vinyals, Andrew Zisserman, and Karen Simonyan.
\newblock Flamingo: a visual language model for few-shot learning.
\newblock \emph{ArXiv}, abs/2204.14198, 2022.

\bibitem[Alfassy et~al.(2022)Alfassy, Arbelle, Halimi, Harary, Herzig, Schwartz, Panda, Dolfi, Auer, Staar, Saenko, Feris, and Karlinsky]{alfassy2022feta}
Amit Alfassy, Assaf Arbelle, Oshri Halimi, Sivan Harary, Roei Herzig, Eli Schwartz, Rameswar Panda, Michele Dolfi, Christoph Auer, Peter W.~J. Staar, Kate Saenko, Rogerio Feris, and Leonid Karlinsky.
\newblock {FETA}: Towards specializing foundational models for expert task applications.
\newblock In \emph{Thirty-sixth Conference on Neural Information Processing Systems Datasets and Benchmarks Track}, 2022.

\bibitem[Antol et~al.(2015)Antol, Agrawal, Lu, Mitchell, Batra, Zitnick, and Parikh]{antol2015vqa}
Stanislaw Antol, Aishwarya Agrawal, Jiasen Lu, Margaret Mitchell, Dhruv Batra, C~Lawrence Zitnick, and Devi Parikh.
\newblock Vqa: Visual question answering.
\newblock In \emph{Proceedings of the IEEE international conference on computer vision}, pages 2425--2433, 2015.

\bibitem[Avraham et~al.(2022)Avraham, Herzig, Mangalam, Bar, Rohrbach, Karlinsky, Darrell, and Globerson]{avraham2022svit}
Elad~Ben Avraham, Roei Herzig, Karttikeya Mangalam, Amir Bar, Anna Rohrbach, Leonid Karlinsky, Trevor Darrell, and Amir Globerson.
\newblock Bringing image scene structure to video via frame-clip consistency of object tokens.
\newblock In \emph{Thirty-Sixth Conference on Neural Information Processing Systems}, 2022.

\bibitem[Bai et~al.(2023)Bai, Bai, Yang, Wang, Tan, Wang, Lin, Zhou, and Zhou]{Bai2023QwenVLAF}
Jinze Bai, Shuai Bai, Shusheng Yang, Shijie Wang, Sinan Tan, Peng Wang, Junyang Lin, Chang Zhou, and Jingren Zhou.
\newblock Qwen-vl: A frontier large vision-language model with versatile abilities.
\newblock \emph{ArXiv}, abs/2308.12966, 2023.

\bibitem[Bertsch et~al.(2024)Bertsch, Ivgi, Alon, Berant, Gormley, and Neubig]{Bertsch2024InContextLW}
Amanda Bertsch, Maor Ivgi, Uri Alon, Jonathan Berant, Matthew~R. Gormley, and Graham Neubig.
\newblock In-context learning with long-context models: An in-depth exploration.
\newblock 2024.

\bibitem[Besta et~al.(2023)Besta, Blach, Kubicek, Gerstenberger, Gianinazzi, Gajda, Lehmann, Podstawski, Niewiadomski, Nyczyk, and Hoefler]{Besta2023GraphOT}
Maciej Besta, Nils Blach, Ales Kubicek, Robert Gerstenberger, Lukas Gianinazzi, Joanna Gajda, Tomasz Lehmann, Michal Podstawski, Hubert Niewiadomski, Piotr Nyczyk, and Torsten Hoefler.
\newblock Graph of thoughts: Solving elaborate problems with large language models.
\newblock \emph{ArXiv}, abs/2308.09687, 2023.

\bibitem[Brown et~al.(2020)Brown, Mann, Ryder, Subbiah, Kaplan, Dhariwal, Neelakantan, Shyam, Sastry, Askell, Agarwal, Herbert-Voss, Krueger, Henighan, Child, Ramesh, Ziegler, Wu, Winter, Hesse, Chen, Sigler, Litwin, Gray, Chess, Clark, Berner, McCandlish, Radford, Sutskever, and Amodei]{Brown2020OGICL}
Tom~B. Brown, Benjamin Mann, Nick Ryder, Melanie Subbiah, Jared Kaplan, Prafulla Dhariwal, Arvind Neelakantan, Pranav Shyam, Girish Sastry, Amanda Askell, Sandhini Agarwal, Ariel Herbert-Voss, Gretchen Krueger, Tom Henighan, Rewon Child, Aditya Ramesh, Daniel~M. Ziegler, Jeff Wu, Clemens Winter, Christopher Hesse, Mark Chen, Eric Sigler, Mateusz Litwin, Scott Gray, Benjamin Chess, Jack Clark, Christopher Berner, Sam McCandlish, Alec Radford, Ilya Sutskever, and Dario Amodei.
\newblock Language models are few-shot learners.
\newblock \emph{ArXiv}, abs/2005.14165, 2020.

\bibitem[Chen et~al.(2023)Chen, Wong, Chen, and Tian]{Chen2023PosInter}
Shouyuan Chen, Sherman Wong, Liangjian Chen, and Yuandong Tian.
\newblock Extending context window of large language models via positional interpolation.
\newblock \emph{ArXiv}, abs/2306.15595, 2023.

\bibitem[Chevalier et~al.(2023)Chevalier, Wettig, Ajith, and Chen]{ChevalierAutoCOmp}
Alexis Chevalier, Alexander Wettig, Anirudh Ajith, and Danqi Chen.
\newblock Adapting language models to compress contexts.
\newblock \emph{ArXiv}, abs/2305.14788, 2023.

\bibitem[Chowdhery et~al.(2022)Chowdhery, Narang, Devlin, Bosma, Mishra, Roberts, Barham, Chung, Sutton, Gehrmann, Schuh, Shi, Tsvyashchenko, Maynez, Rao, Barnes, Tay, Shazeer, Prabhakaran, Reif, Du, Hutchinson, Pope, Bradbury, Austin, Isard, Gur-Ari, Yin, Duke, Levskaya, Ghemawat, Dev, Michalewski, Garc{\'i}a, Misra, Robinson, Fedus, Zhou, Ippolito, Luan, Lim, Zoph, Spiridonov, Sepassi, Dohan, Agrawal, Omernick, Dai, Pillai, Pellat, Lewkowycz, Moreira, Child, Polozov, Lee, Zhou, Wang, Saeta, D{\'i}az, Firat, Catasta, Wei, Meier-Hellstern, Eck, Dean, Petrov, and Fiedel]{Chowdhery2022PaLMSL}
Aakanksha Chowdhery, Sharan Narang, Jacob Devlin, Maarten Bosma, Gaurav Mishra, Adam Roberts, Paul Barham, Hyung~Won Chung, Charles Sutton, Sebastian Gehrmann, Parker Schuh, Kensen Shi, Sasha Tsvyashchenko, Joshua Maynez, Abhishek Rao, Parker Barnes, Yi Tay, Noam~M. Shazeer, Vinodkumar Prabhakaran, Emily Reif, Nan Du, Benton~C. Hutchinson, Reiner Pope, James Bradbury, Jacob Austin, Michael Isard, Guy Gur-Ari, Pengcheng Yin, Toju Duke, Anselm Levskaya, Sanjay Ghemawat, Sunipa Dev, Henryk Michalewski, Xavier Garc{\'i}a, Vedant Misra, Kevin Robinson, Liam Fedus, Denny Zhou, Daphne Ippolito, David Luan, Hyeontaek Lim, Barret Zoph, Alexander Spiridonov, Ryan Sepassi, David Dohan, Shivani Agrawal, Mark Omernick, Andrew~M. Dai, Thanumalayan~Sankaranarayana Pillai, Marie Pellat, Aitor Lewkowycz, Erica Moreira, Rewon Child, Oleksandr Polozov, Katherine Lee, Zongwei Zhou, Xuezhi Wang, Brennan Saeta, Mark D{\'i}az, Orhan Firat, Michele Catasta, Jason Wei, Kathleen~S. Meier-Hellstern, Douglas Eck, Jeff Dean, Slav Petrov,
  and Noah Fiedel.
\newblock Palm: Scaling language modeling with pathways.
\newblock \emph{J. Mach. Learn. Res.}, 24:\penalty0 240:1--240:113, 2022.

\bibitem[Dai et~al.(2023)Dai, Li, Li, Tiong, Zhao, Wang, Li, Fung, and Hoi]{instructblip}
Wenliang Dai, Junnan Li, Dongxu Li, Anthony Meng~Huat Tiong, Junqi Zhao, Weisheng Wang, Boyang Li, Pascale Fung, and Steven Hoi.
\newblock Instructblip: Towards general-purpose vision-language models with instruction tuning, 2023.

\bibitem[Dettmers et~al.(2023)Dettmers, Pagnoni, Holtzman, and Zettlemoyer]{Dettmers2023QLoRAEF}
Tim Dettmers, Artidoro Pagnoni, Ari Holtzman, and Luke Zettlemoyer.
\newblock Qlora: Efficient finetuning of quantized llms.
\newblock \emph{ArXiv}, abs/2305.14314, 2023.

\bibitem[Dong et~al.(2022)Dong, Li, Dai, Zheng, Wu, Chang, Sun, Xu, and Sui]{Dong2022ASO}
Qingxiu Dong, Lei Li, Damai Dai, Ce Zheng, Zhiyong Wu, Baobao Chang, Xu Sun, Jingjing Xu, and Zhifang Sui.
\newblock A survey on in-context learning.
\newblock 2022.

\bibitem[Doveh et~al.(2024)Doveh, Perek, Mirza, Alfassy, Arbelle, Ullman, and Karlinsky]{Doveh2024TowardsMM-ICL}
Sivan Doveh, Shaked Perek, Muhammad~Jehanzeb Mirza, Amit Alfassy, Assaf Arbelle, Shimon Ullman, and Leonid Karlinsky.
\newblock Towards multimodal in-context learning for vision \& language models.
\newblock \emph{ArXiv}, abs/2403.12736, 2024.

\bibitem[Driess et~al.(2023)Driess, Xia, Sajjadi, Lynch, Chowdhery, Ichter, Wahid, Tompson, Vuong, Yu, Huang, Chebotar, Sermanet, Duckworth, Levine, Vanhoucke, Hausman, Toussaint, Greff, Zeng, Mordatch, and Florence]{Driess2023PaLMEAE}
Danny Driess, F. Xia, Mehdi S.~M. Sajjadi, Corey Lynch, Aakanksha Chowdhery, Brian Ichter, Ayzaan Wahid, Jonathan Tompson, Quan~Ho Vuong, Tianhe Yu, Wenlong Huang, Yevgen Chebotar, Pierre Sermanet, Daniel Duckworth, Sergey Levine, Vincent Vanhoucke, Karol Hausman, Marc Toussaint, Klaus Greff, Andy Zeng, Igor Mordatch, and Peter~R. Florence.
\newblock Palm-e: An embodied multimodal language model.
\newblock In \emph{International Conference on Machine Learning}, 2023.

\bibitem[Gao et~al.(2023)Gao, Han, Zhang, Lin, Geng, Zhou, Zhang, Lu, He, Yue, Li, and Qiao]{Gao2023LLaMAAdapter2}
Peng Gao, Jiaming Han, Renrui Zhang, Ziyi Lin, Shijie Geng, Aojun Zhou, W. Zhang, Pan Lu, Conghui He, Xiangyu Yue, Hongsheng Li, and Yu~Jiao Qiao.
\newblock Llama-adapter v2: Parameter-efficient visual instruction model.
\newblock \emph{ArXiv}, abs/2304.15010, 2023.

\bibitem[Ge et~al.(2025)Ge, Subramanian, Shi, Herzig, and Darrell]{Ge2023RecursiveVP}
Jiaxin Ge, Sanjay Subramanian, Baifeng Shi, Roei Herzig, and Trevor Darrell.
\newblock Recursive visual programming.
\newblock In \emph{European Conference on Computer Vision}, pages 1--18. Springer, 2025.

\bibitem[Ge et~al.(2023)Ge, Hu, Wang, Chen, and Wei]{Ge2023IncontextAF}
Tao Ge, Jing Hu, Xun Wang, Si-Qing Chen, and Furu Wei.
\newblock In-context autoencoder for context compression in a large language model.
\newblock \emph{ArXiv}, abs/2307.06945, 2023.

\bibitem[Gong et~al.(2023)Gong, Lyu, Zhang, Wang, Zheng, Zhao, Liu, Zhang, Luo, and Chen]{Gong2023MultiModalGPTAV}
Tao Gong, Chengqi Lyu, Shilong Zhang, Yudong Wang, Miao Zheng, Qianmengke Zhao, Kuikun Liu, Wenwei Zhang, Ping Luo, and Kai Chen.
\newblock Multimodal-gpt: A vision and language model for dialogue with humans.
\newblock \emph{ArXiv}, abs/2305.04790, 2023.

\bibitem[Gupta and Kembhavi(2022)]{Gupta2022VisualPC}
Tanmay Gupta and Aniruddha Kembhavi.
\newblock Visual programming: Compositional visual reasoning without training.
\newblock \emph{2023 IEEE/CVF Conference on Computer Vision and Pattern Recognition (CVPR)}, pages 14953--14962, 2022.

\bibitem[Gurari et~al.(2018)Gurari, Li, Stangl, Guo, Lin, Grauman, Luo, and Bigham]{Gurari2018VizWizGC}
Danna Gurari, Qing Li, Abigale Stangl, Anhong Guo, Chi Lin, Kristen Grauman, Jiebo Luo, and Jeffrey~P. Bigham.
\newblock Vizwiz grand challenge: Answering visual questions from blind people.
\newblock \emph{2018 IEEE/CVF Conference on Computer Vision and Pattern Recognition}, pages 3608--3617, 2018.

\bibitem[He et~al.(2019)He, Fan, Wu, Xie, and Girshick]{He2019MomentumCF}
Kaiming He, Haoqi Fan, Yuxin Wu, Saining Xie, and Ross~B. Girshick.
\newblock Momentum contrast for unsupervised visual representation learning.
\newblock \emph{2020 IEEE/CVF Conference on Computer Vision and Pattern Recognition (CVPR)}, pages 9726--9735, 2019.

\bibitem[Hendel et~al.(2023)Hendel, Geva, and Globerson]{Hendel2023InContextLC}
Roee Hendel, Mor Geva, and Amir Globerson.
\newblock In-context learning creates task vectors.
\newblock \emph{ArXiv}, abs/2310.15916, 2023.

\bibitem[Herzig et~al.(2023)Herzig, Mendelson, Karlinsky, Arbelle, Feris, Darrell, and Globerson]{herzig2023incorporating}
Roei Herzig, Alon Mendelson, Leonid Karlinsky, Assaf Arbelle, Rogerio Feris, Trevor Darrell, and Amir Globerson.
\newblock Incorporating structured representations into pretrained vision {\textbackslash}\& language models using scene graphs.
\newblock In \emph{The 2023 Conference on Empirical Methods in Natural Language Processing}, 2023.

\bibitem[Hojel et~al.(2025)Hojel, Bai, Darrell, Globerson, and Bar]{Hojel2024FindingVT}
Alberto Hojel, Yutong Bai, Trevor Darrell, Amir Globerson, and Amir Bar.
\newblock Finding visual task vectors.
\newblock In \emph{European Conference on Computer Vision}, pages 257--273. Springer, 2025.

\bibitem[Houlsby et~al.(2019)Houlsby, Giurgiu, Jastrzebski, Morrone, De~Laroussilhe, Gesmundo, Attariyan, and Gelly]{houlsby19adapters}
Neil Houlsby, Andrei Giurgiu, Stanislaw Jastrzebski, Bruna Morrone, Quentin De~Laroussilhe, Andrea Gesmundo, Mona Attariyan, and Sylvain Gelly.
\newblock Parameter-efficient transfer learning for {NLP}.
\newblock In \emph{Proceedings of the 36th International Conference on Machine Learning}, pages 2790--2799, 2019.

\bibitem[Hu et~al.(2021)Hu, Shen, Wallis, Allen-Zhu, Li, Wang, and Chen]{Hu2021LoRALA}
J.~Edward Hu, Yelong Shen, Phillip Wallis, Zeyuan Allen-Zhu, Yuanzhi Li, Shean Wang, and Weizhu Chen.
\newblock Lora: Low-rank adaptation of large language models.
\newblock \emph{ArXiv}, abs/2106.09685, 2021.

\bibitem[Hu et~al.(2023)Hu, Lan, Wang, Xu, Lim, Lee, Bing, and Poria]{Hu2023LLMAdaptersAA}
Zhiqiang Hu, Yihuai Lan, Lei Wang, Wanyu Xu, Ee-Peng Lim, Roy Ka-Wei Lee, Lidong Bing, and Soujanya Poria.
\newblock Llm-adapters: An adapter family for parameter-efficient fine-tuning of large language models.
\newblock \emph{ArXiv}, abs/2304.01933, 2023.

\bibitem[Hudson and Manning(2019)]{Hudson2019GQAAN}
Drew~A. Hudson and Christopher~D. Manning.
\newblock Gqa: A new dataset for real-world visual reasoning and compositional question answering.
\newblock \emph{2019 IEEE/CVF Conference on Computer Vision and Pattern Recognition (CVPR)}, pages 6693--6702, 2019.

\bibitem[Jia et~al.(2021)Jia, Yang, Xia, Chen, Parekh, Pham, Le, Sung, Li, and Duerig]{Jia2021ScalingUV}
Chao Jia, Yinfei Yang, Ye Xia, Yi-Ting Chen, Zarana Parekh, Hieu Pham, Quoc~V. Le, Yun-Hsuan Sung, Zhen Li, and Tom Duerig.
\newblock Scaling up visual and vision-language representation learning with noisy text supervision.
\newblock In \emph{International Conference on Machine Learning}, 2021.

\bibitem[Jiang et~al.(2024)Jiang, He, Zeng, Wei, Ku, Liu, and Chen]{Jiang2024MANTISIM}
Dongfu Jiang, Xuan He, Huaye Zeng, Cong Wei, Max~W.F. Ku, Qian Liu, and Wenhu Chen.
\newblock Mantis: Interleaved multi-image instruction tuning.
\newblock arXiv2405.01483, 2024.

\bibitem[Jiang et~al.(2023)Jiang, Wu, Lin, Yang, and Qiu]{Jiang2023LLMLinguaCP}
Huiqiang Jiang, Qianhui Wu, Chin-Yew Lin, Yuqing Yang, and Lili Qiu.
\newblock Llmlingua: Compressing prompts for accelerated inference of large language models.
\newblock In \emph{Conference on Empirical Methods in Natural Language Processing}, 2023.

\bibitem[Jiang et~al.()Jiang, Irvin, Wang, Chaudhry, Chen, and Ng]{jiang2024many}
Yixing Jiang, Jeremy~Andrew Irvin, Ji~Hun Wang, Muhammad~Ahmed Chaudhry, Jonathan~H Chen, and Andrew~Y Ng.
\newblock Many-shot in-context learning in multimodal foundation models.
\newblock In \emph{ICML 2024 Workshop on In-Context Learning}.

\bibitem[Kojima et~al.(2022)Kojima, Gu, Reid, Matsuo, and Iwasawa]{Kojima2022LargeLM}
Takeshi Kojima, Shixiang~Shane Gu, Machel Reid, Yutaka Matsuo, and Yusuke Iwasawa.
\newblock Large language models are zero-shot reasoners.
\newblock \emph{ArXiv}, abs/2205.11916, 2022.

\bibitem[Krishna et~al.(2017)Krishna, Zhu, Groth, Johnson, Hata, Kravitz, Chen, Kalantidis, Li, Shamma, et~al.]{krishna2017visual}
Ranjay Krishna, Yuke Zhu, Oliver Groth, Justin Johnson, Kenji Hata, Joshua Kravitz, Stephanie Chen, Yannis Kalantidis, Li-Jia Li, David~A Shamma, et~al.
\newblock Visual genome: Connecting language and vision using crowdsourced dense image annotations.
\newblock \emph{International Journal of Computer Vision}, 123\penalty0 (1):\penalty0 32--73, 2017.

\bibitem[Laurenccon et~al.(2023)Laurenccon, Saulnier, Tronchon, Bekman, Singh, Lozhkov, Wang, Karamcheti, Rush, Kiela, Cord, and Sanh]{Laurenccon2023IDEFICS}
Hugo Laurenccon, Lucile Saulnier, L{\'e}o Tronchon, Stas Bekman, Amanpreet Singh, Anton Lozhkov, Thomas Wang, Siddharth Karamcheti, Alexander~M. Rush, Douwe Kiela, Matthieu Cord, and Victor Sanh.
\newblock Obelisc: An open web-scale filtered dataset of interleaved image-text documents.
\newblock \emph{ArXiv}, abs/2306.16527, 2023.

\bibitem[Laurenccon et~al.(2024)Laurenccon, Tronchon, Cord, and Sanh]{Laurenccon2024Idefics2}
Hugo Laurenccon, L{\'e}o Tronchon, Matthieu Cord, and Victor Sanh.
\newblock What matters when building vision-language models?
\newblock 2024.

\bibitem[Lei et~al.(2023)Lei, Lin, Liao, and Ding]{Lei2023BoostingLR}
Bin Lei, Pei-Hung Lin, Chunhua Liao, and Caiwen Ding.
\newblock Boosting logical reasoning in large language models through a new framework: The graph of thought.
\newblock \emph{ArXiv}, abs/2308.08614, 2023.

\bibitem[Lester et~al.(2021)Lester, Al-Rfou, and Constant]{Lester2021ThePO}
Brian Lester, Rami Al-Rfou, and Noah Constant.
\newblock The power of scale for parameter-efficient prompt tuning.
\newblock In \emph{Conference on Empirical Methods in Natural Language Processing}, 2021.

\bibitem[Li et~al.(2022)Li, Li, Xiong, and Hoi]{blip}
Junnan Li, Dongxu Li, Caiming Xiong, and Steven Hoi.
\newblock Blip: Bootstrapping language-image pre-training for unified vision-language understanding and generation.
\newblock \emph{arXiv preprint arXiv:2201.12086}, 2022.

\bibitem[Li et~al.(2023{\natexlab{a}})Li, Li, Savarese, and Hoi]{li2023blip2}
Junnan Li, Dongxu Li, Silvio Savarese, and Steven Hoi.
\newblock {BLIP-2:} bootstrapping language-image pre-training with frozen image encoders and large language models.
\newblock In \emph{ICML}, 2023{\natexlab{a}}.

\bibitem[Li et~al.(2023{\natexlab{b}})Li, Gong, Feng, Xu, Zhang, Wu, and Kong]{Li2023InContextLW}
Mukai Li, Shansan Gong, Jiangtao Feng, Yiheng Xu, Jinchao Zhang, Zhiyong Wu, and Lingpeng Kong.
\newblock In-context learning with many demonstration examples.
\newblock \emph{ArXiv}, abs/2302.04931, 2023{\natexlab{b}}.

\bibitem[Li et~al.(2024)Li, Zhang, Do, Yue, and Chen]{Li2024LongcontextLS}
Tianle Li, Ge Zhang, Quy~Duc Do, Xiang Yue, and Wenhu Chen.
\newblock Long-context llms struggle with long in-context learning.
\newblock 2024.

\bibitem[Li and Liang(2021)]{Li2021PrefixTuningOC}
Xiang~Lisa Li and Percy Liang.
\newblock Prefix-tuning: Optimizing continuous prompts for generation.
\newblock \emph{Proceedings of the 59th Annual Meeting of the Association for Computational Linguistics and the 11th International Joint Conference on Natural Language Processing (Volume 1: Long Papers)}, abs/2101.00190, 2021.

\bibitem[Lin et~al.(2023)Lin, Yin, Ping, Lu, Molchanov, Tao, Mao, Kautz, Shoeybi, and Han]{lin2023vila}
Ji Lin, Hongxu Yin, Wei Ping, Yao Lu, Pavlo Molchanov, Andrew Tao, Huizi Mao, Jan Kautz, Mohammad Shoeybi, and Song Han.
\newblock Vila: On pre-training for visual language models, 2023.

\bibitem[Lin et~al.(2014)Lin, Maire, Belongie, Hays, Perona, Ramanan, Doll{\'a}r, and Zitnick]{Lin2014MSCOCO}
Tsung-Yi Lin, M. Maire, Serge~J. Belongie, James Hays, P. Perona, D. Ramanan, Piotr Doll{\'a}r, and C.~L. Zitnick.
\newblock Microsoft coco: Common objects in context.
\newblock In \emph{ECCV}, 2014.

\bibitem[Liu et~al.(2023{\natexlab{a}})Liu, Li, Li, and Lee]{liu2023llava15}
Haotian Liu, Chunyuan Li, Yuheng Li, and Yong~Jae Lee.
\newblock Improved baselines with visual instruction tuning, 2023{\natexlab{a}}.

\bibitem[Liu et~al.(2023{\natexlab{b}})Liu, Li, Wu, and Lee]{liu2023llava}
Haotian Liu, Chunyuan Li, Qingyang Wu, and Yong~Jae Lee.
\newblock Visual instruction tuning.
\newblock In \emph{NeurIPS}, 2023{\natexlab{b}}.

\bibitem[Liu et~al.(2023{\natexlab{c}})Liu, Lin, Hewitt, Paranjape, Bevilacqua, Petroni, and Liang]{Liu2023LostIT}
Nelson~F. Liu, Kevin Lin, John Hewitt, Ashwin Paranjape, Michele Bevilacqua, Fabio Petroni, and Percy Liang.
\newblock Lost in the middle: How language models use long contexts.
\newblock \emph{Transactions of the Association for Computational Linguistics}, 12:\penalty0 157--173, 2023{\natexlab{c}}.

\bibitem[Lu et~al.(2023)Lu, Peng, Cheng, Galley, Chang, Wu, Zhu, and Gao]{Lu2023ChameleonPC}
Pan Lu, Baolin Peng, Hao Cheng, Michel Galley, Kai-Wei Chang, Ying~Nian Wu, Song-Chun Zhu, and Jianfeng Gao.
\newblock Chameleon: Plug-and-play compositional reasoning with large language models.
\newblock \emph{ArXiv}, abs/2304.09842, 2023.

\bibitem[Ma et~al.(2023)Ma, Zhang, Bian, Liu, Zhang, Zhao, Zhang, Fu, Hu, and Wu]{Ma2023FairnessguidedFP}
Huan Ma, Changqing Zhang, Yatao Bian, Lemao Liu, Zhirui Zhang, Peilin Zhao, Shu Zhang, H. Fu, Qinghua Hu, and Bing Wu.
\newblock Fairness-guided few-shot prompting for large language models.
\newblock \emph{ArXiv}, abs/2303.13217, 2023.

\bibitem[Marino et~al.(2019)Marino, Rastegari, Farhadi, and Mottaghi]{Marino2019OKVQAAV}
Kenneth Marino, Mohammad Rastegari, Ali Farhadi, and Roozbeh Mottaghi.
\newblock Ok-vqa: A visual question answering benchmark requiring external knowledge.
\newblock \emph{2019 IEEE/CVF Conference on Computer Vision and Pattern Recognition (CVPR)}, pages 3190--3199, 2019.

\bibitem[Meta et~al.(2024)Meta, Jauhri, Pandey, Kadian, Al-Dahle, Letman, Mathur, Schelten, Yang, Fan, et~al.]{meta2024llama3}
AI Meta, Abhinav Jauhri, Abhinav Pandey, Abhishek Kadian, Ahmad Al-Dahle, Aiesha Letman, Akhil Mathur, Alan Schelten, Amy Yang, Angela Fan, et~al.
\newblock The llama 3 herd of models.
\newblock \emph{arXiv preprint arXiv:2407.21783}, 2, 2024.

\bibitem[Min et~al.(2022)Min, Lyu, Holtzman, Artetxe, Lewis, Hajishirzi, and Zettlemoyer]{Min2022RethinkingTR}
Sewon Min, Xinxi Lyu, Ari Holtzman, Mikel Artetxe, Mike Lewis, Hannaneh Hajishirzi, and Luke Zettlemoyer.
\newblock Rethinking the role of demonstrations: What makes in-context learning work?
\newblock \emph{ArXiv}, abs/2202.12837, 2022.

\bibitem[Mitra et~al.(2024)Mitra, Huang, Darrell, and Herzig]{MitraCCoT}
Chancharik Mitra, Brandon Huang, Trevor Darrell, and Roei Herzig.
\newblock Compositional chain of thought prompting for large multimodal models.
\newblock In \emph{Proceedings of the IEEE/CVF Conference on Computer Vision and Pattern Recognition (CVPR)}, 2024.

\bibitem[Mu et~al.(2023)Mu, Li, and Goodman]{Mu2023LearningTC}
Jesse Mu, Xiang~Lisa Li, and Noah~D. Goodman.
\newblock Learning to compress prompts with gist tokens.
\newblock \emph{ArXiv}, abs/2304.08467, 2023.

\bibitem[Nilsback and Zisserman(2008)]{Flowers}
Maria-Elena Nilsback and Andrew Zisserman.
\newblock Automated flower classification over a large number of classes.
\newblock \emph{2008 Sixth Indian Conference on Computer Vision, Graphics \& Image Processing}, pages 722--729, 2008.

\bibitem[OpenAI(2023)]{OpenAI2023GPT4TR}
OpenAI.
\newblock Gpt-4 technical report.
\newblock \emph{ArXiv}, abs/2303.08774, 2023.

\bibitem[Paszke et~al.(2019)Paszke, Gross, Massa, Lerer, Bradbury, Chanan, Killeen, Lin, Gimelshein, Antiga, et~al.]{paszke2019pytorch}
Adam Paszke, Sam Gross, Francisco Massa, Adam Lerer, James Bradbury, Gregory Chanan, Trevor Killeen, Zeming Lin, Natalia Gimelshein, Luca Antiga, et~al.
\newblock Pytorch: An imperative style, high-performance deep learning library.
\newblock \emph{Advances in neural information processing systems}, 32, 2019.

\bibitem[Pearl(2001)]{Pearl2001DirectAI}
Judea Pearl.
\newblock Direct and indirect effects.
\newblock \emph{Probabilistic and Causal Inference}, 2001.

\bibitem[Peng et~al.(2023)Peng, Quesnelle, Fan, and Shippole]{Peng2023YaRNEC}
Bowen Peng, Jeffrey Quesnelle, Honglu Fan, and Enrico Shippole.
\newblock Yarn: Efficient context window extension of large language models.
\newblock \emph{ArXiv}, abs/2309.00071, 2023.

\bibitem[Qin et~al.(2023)Qin, Liang, Ye, Zhu, Yan, Lu, Lin, Cong, Tang, Qian, Zhao, Tian, Xie, Zhou, Gerstein, Li, Liu, and Sun]{Qin2023ToolLLMFL}
Yujia Qin, Shi Liang, Yining Ye, Kunlun Zhu, Lan Yan, Ya-Ting Lu, Yankai Lin, Xin Cong, Xiangru Tang, Bill Qian, Sihan Zhao, Runchu Tian, Ruobing Xie, Jie Zhou, Marc~H. Gerstein, Dahai Li, Zhiyuan Liu, and Maosong Sun.
\newblock Toolllm: Facilitating large language models to master 16000+ real-world apis.
\newblock \emph{ArXiv}, abs/2307.16789, 2023.

\bibitem[Radford et~al.(2021)Radford, Kim, Hallacy, Ramesh, Goh, Agarwal, Sastry, Askell, Mishkin, Clark, et~al.]{radford2021clip}
Alec Radford, Jong~Wook Kim, Chris Hallacy, Aditya Ramesh, Gabriel Goh, Sandhini Agarwal, Girish Sastry, Amanda Askell, Pamela Mishkin, Jack Clark, et~al.
\newblock Learning transferable visual models from natural language supervision.
\newblock In \emph{International Conference on Machine Learning}, pages 8748--8763. PMLR, 2021.

\bibitem[Raffel et~al.(2019)Raffel, Shazeer, Roberts, Lee, Narang, Matena, Zhou, Li, and Liu]{Raffel2019ExploringTL}
Colin Raffel, Noam~M. Shazeer, Adam Roberts, Katherine Lee, Sharan Narang, Michael Matena, Yanqi Zhou, Wei Li, and Peter~J. Liu.
\newblock Exploring the limits of transfer learning with a unified text-to-text transformer.
\newblock \emph{J. Mach. Learn. Res.}, 21:\penalty0 140:1--140:67, 2019.

\bibitem[Saikh et~al.(2022)Saikh, Ghosal, Mittal, Ekbal, and Bhattacharyya]{Saikh2022ScienceQAAN}
Tanik Saikh, Tirthankar Ghosal, Amish Mittal, Asif Ekbal, and Pushpak Bhattacharyya.
\newblock Scienceqa: a novel resource for question answering on scholarly articles.
\newblock \emph{International Journal on Digital Libraries}, 23:\penalty0 289 -- 301, 2022.

\bibitem[Sanh et~al.(2022)Sanh, Webson, Raffel, Bach, Sutawika, Alyafeai, Chaffin, Stiegler, Raja, Dey, Bari, Xu, Thakker, Sharma, Szczechla, Kim, Chhablani, Nayak, Datta, Chang, Jiang, Wang, Manica, Shen, Yong, Pandey, Bawden, Wang, Neeraj, Rozen, Sharma, Santilli, Fevry, Fries, Teehan, Scao, Biderman, Gao, Wolf, and Rush]{sanh2022multitask}
Victor Sanh, Albert Webson, Colin Raffel, Stephen Bach, Lintang Sutawika, Zaid Alyafeai, Antoine Chaffin, Arnaud Stiegler, Arun Raja, Manan Dey, M~Saiful Bari, Canwen Xu, Urmish Thakker, Shanya~Sharma Sharma, Eliza Szczechla, Taewoon Kim, Gunjan Chhablani, Nihal Nayak, Debajyoti Datta, Jonathan Chang, Mike Tian-Jian Jiang, Han Wang, Matteo Manica, Sheng Shen, Zheng~Xin Yong, Harshit Pandey, Rachel Bawden, Thomas Wang, Trishala Neeraj, Jos Rozen, Abheesht Sharma, Andrea Santilli, Thibault Fevry, Jason~Alan Fries, Ryan Teehan, Teven~Le Scao, Stella Biderman, Leo Gao, Thomas Wolf, and Alexander~M Rush.
\newblock Multitask prompted training enables zero-shot task generalization.
\newblock In \emph{International Conference on Learning Representations}, 2022.

\bibitem[Schick et~al.(2023)Schick, Dwivedi-Yu, Dess{\`i}, Raileanu, Lomeli, Zettlemoyer, Cancedda, and Scialom]{Schick2023ToolformerLM}
Timo Schick, Jane Dwivedi-Yu, Roberto Dess{\`i}, Roberta Raileanu, Maria Lomeli, Luke Zettlemoyer, Nicola Cancedda, and Thomas Scialom.
\newblock Toolformer: Language models can teach themselves to use tools.
\newblock \emph{ArXiv}, abs/2302.04761, 2023.

\bibitem[Schreiber et~al.(2020)Schreiber, Bilmes, and Noble]{schreiber2020apricot}
Jacob Schreiber, Jeffrey Bilmes, and William~Stafford Noble.
\newblock apricot: Submodular selection for data summarization in python.
\newblock \emph{Journal of Machine Learning Research}, 21\penalty0 (161):\penalty0 1--6, 2020.

\bibitem[Shang et~al.(2024)Shang, You, Subramanian, Darrell, and Herzig]{Shang2024TraveLERAM}
Chuyi Shang, Amos You, Sanjay Subramanian, Trevor Darrell, and Roei Herzig.
\newblock Traveler: A multi-lmm agent framework for video question-answering.
\newblock \emph{ArXiv}, abs/2404.01476, 2024.

\bibitem[Shen et~al.(2023)Shen, Song, Tan, Li, Lu, and Zhuang]{Shen2023HuggingGPTSA}
Yongliang Shen, Kaitao Song, Xu Tan, Dong~Sheng Li, Weiming Lu, and Yue~Ting Zhuang.
\newblock Hugginggpt: Solving ai tasks with chatgpt and its friends in hugging face.
\newblock \emph{ArXiv}, abs/2303.17580, 2023.

\bibitem[Snell et~al.(2022)Snell, Klein, and Zhong]{Snell2022LearningBD}
Charles~Burton Snell, Dan Klein, and Ruiqi Zhong.
\newblock Learning by distilling context.
\newblock \emph{ArXiv}, abs/2209.15189, 2022.

\bibitem[Subramanian et~al.(2023)Subramanian, Narasimhan, Khangaonkar, Yang, Nagrani, Schmid, Zeng, Darrell, and Klein]{Subramanian2023ModularVQ}
Sanjay Subramanian, Medhini~G. Narasimhan, Kushal Khangaonkar, Kevin Yang, Arsha Nagrani, Cordelia Schmid, Andy Zeng, Trevor Darrell, and Dan Klein.
\newblock Modular visual question answering via code generation.
\newblock \emph{ArXiv}, abs/2306.05392, 2023.

\bibitem[Sun et~al.(2023)Sun, Cui, Zhang, Zhang, Yu, Luo, Wang, Rao, Liu, Huang, and Wang]{Sun2023Emu2}
Quan Sun, Yufeng Cui, Xiaosong Zhang, Fan Zhang, Qiying Yu, Zhengxiong Luo, Yueze Wang, Yongming Rao, Jingjing Liu, Tiejun Huang, and Xinlong Wang.
\newblock Generative multimodal models are in-context learners.
\newblock \emph{ArXiv}, abs/2312.13286, 2023.

\bibitem[Sur'is et~al.(2023)Sur'is, Menon, and Vondrick]{Suris2023ViperGPTVI}
D'idac Sur'is, Sachit Menon, and Carl Vondrick.
\newblock Vipergpt: Visual inference via python execution for reasoning.
\newblock \emph{ArXiv}, abs/2303.08128, 2023.

\bibitem[Tan et~al.(2024)Tan, Li, Patil, Wu, Zhang, Keutzer, Gonzalez, and Popa]{Tan2024LLoCOLL}
Sijun Tan, Xiuyu Li, Shishir~G. Patil, Ziyang Wu, Tianjun Zhang, Kurt Keutzer, Joseph~E. Gonzalez, and Raluca~A. Popa.
\newblock Lloco: Learning long contexts offline.
\newblock 2024.

\bibitem[Tay et~al.(2022)Tay, Dehghani, Tran, Garc{\'i}a, Wei, Wang, Chung, Bahri, Schuster, Zheng, Zhou, Houlsby, and Metzler]{Tay2022UL2UL}
Yi Tay, Mostafa Dehghani, Vinh~Q. Tran, Xavier Garc{\'i}a, Jason Wei, Xuezhi Wang, Hyung~Won Chung, Dara Bahri, Tal Schuster, Huaixiu~Steven Zheng, Denny Zhou, Neil Houlsby, and Donald Metzler.
\newblock Ul2: Unifying language learning paradigms.
\newblock In \emph{International Conference on Learning Representations}, 2022.

\bibitem[Team(2024)]{Reid2024Gemini1.5}
Gemini Team.
\newblock Gemini 1.5: Unlocking multimodal understanding across millions of tokens of context.
\newblock \emph{ArXiv}, abs/2403.05530, 2024.

\bibitem[Team et~al.(2023)Team, Anil, Borgeaud, Wu, Alayrac, Yu, Soricut, Schalkwyk, Dai, Hauth, et~al.]{team2023gemini}
Gemini Team, Rohan Anil, Sebastian Borgeaud, Yonghui Wu, Jean-Baptiste Alayrac, Jiahui Yu, Radu Soricut, Johan Schalkwyk, Andrew~M Dai, Anja Hauth, et~al.
\newblock Gemini: a family of highly capable multimodal models.
\newblock \emph{arXiv preprint arXiv:2312.11805}, 2023.

\bibitem[Todd et~al.(2023)Todd, Li, Sharma, Mueller, Wallace, and Bau]{Todd2023FunctionVI}
Eric Todd, Millicent Li, Arnab~Sen Sharma, Aaron Mueller, Byron~C. Wallace, and David Bau.
\newblock Function vectors in large language models.
\newblock \emph{ArXiv}, abs/2310.15213, 2023.

\bibitem[Wah et~al.(2011)Wah, Branson, Welinder, Perona, and Belongie]{Birds}
Catherine Wah, Steve Branson, Peter Welinder, Pietro Perona, and Serge~J. Belongie.
\newblock The caltech-ucsd birds-200-2011 dataset.
\newblock 2011.

\bibitem[Wan et~al.(2023)Wan, Sun, Dai, Arik, and Pfister]{Wan2023BetterZR}
Xingchen Wan, Ruoxi Sun, Hanjun Dai, Sercan~{\"O}. Arik, and Tomas Pfister.
\newblock Better zero-shot reasoning with self-adaptive prompting.
\newblock In \emph{Annual Meeting of the Association for Computational Linguistics}, 2023.

\bibitem[Wang et~al.(2023{\natexlab{a}})Wang, Hu, He, Xu, Liu, juan Liu, and Shen]{Wang2023TSciQTM}
Lei Wang, Yilang Hu, Jiabang He, Xingdong Xu, Ning Liu, Hui juan Liu, and Hengtao Shen.
\newblock T-sciq: Teaching multimodal chain-of-thought reasoning via large language model signals for science question answering.
\newblock \emph{ArXiv}, abs/2305.03453, 2023{\natexlab{a}}.

\bibitem[Wang et~al.(2023{\natexlab{b}})Wang, Xu, Lan, Hu, Lan, Lee, and Lim]{Wang2023PlanandSolvePI}
Lei Wang, Wanyu Xu, Yihuai Lan, Zhiqiang Hu, Yunshi Lan, Roy Ka-Wei Lee, and Ee-Peng Lim.
\newblock Plan-and-solve prompting: Improving zero-shot chain-of-thought reasoning by large language models.
\newblock In \emph{Annual Meeting of the Association for Computational Linguistics}, 2023{\natexlab{b}}.

\bibitem[Wang et~al.(2022{\natexlab{a}})Wang, Wei, Schuurmans, Le, hsin Chi, and Zhou]{Wang2022SelfConsistencyIC}
Xuezhi Wang, Jason Wei, Dale Schuurmans, Quoc Le, Ed~Huai hsin Chi, and Denny Zhou.
\newblock Self-consistency improves chain of thought reasoning in language models.
\newblock \emph{ArXiv}, abs/2203.11171, 2022{\natexlab{a}}.

\bibitem[Wang et~al.(2022{\natexlab{b}})Wang, Li, Xu, Zhou, Lei, Lin, Wang, Yang, Zhu, Hoiem, Chang, Bansal, and Ji]{Wang2022LanguageMW}
Zhenhailong Wang, Manling Li, Ruochen Xu, Luowei Zhou, Jie Lei, Xudong Lin, Shuohang Wang, Ziyi Yang, Chenguang Zhu, Derek Hoiem, Shih-Fu Chang, Mohit Bansal, and Heng Ji.
\newblock Language models with image descriptors are strong few-shot video-language learners.
\newblock \emph{ArXiv}, abs/2205.10747, 2022{\natexlab{b}}.

\bibitem[Wei et~al.(2021)Wei, Bosma, Zhao, Guu, Yu, Lester, Du, Dai, and Le]{Wei2021FinetunedLM}
Jason Wei, Maarten Bosma, Vincent Zhao, Kelvin Guu, Adams~Wei Yu, Brian Lester, Nan Du, Andrew~M. Dai, and Quoc~V. Le.
\newblock Finetuned language models are zero-shot learners.
\newblock \emph{ArXiv}, abs/2109.01652, 2021.

\bibitem[Wei et~al.(2022{\natexlab{a}})Wei, Tay, Bommasani, Raffel, Zoph, Borgeaud, Yogatama, Bosma, Zhou, Metzler, hsin Chi, Hashimoto, Vinyals, Liang, Dean, and Fedus]{Wei2022EmergentAO}
Jason Wei, Yi Tay, Rishi Bommasani, Colin Raffel, Barret Zoph, Sebastian Borgeaud, Dani Yogatama, Maarten Bosma, Denny Zhou, Donald Metzler, Ed~Huai hsin Chi, Tatsunori Hashimoto, Oriol Vinyals, Percy Liang, Jeff Dean, and William Fedus.
\newblock Emergent abilities of large language models.
\newblock \emph{Trans. Mach. Learn. Res.}, 2022, 2022{\natexlab{a}}.

\bibitem[Wei et~al.(2022{\natexlab{b}})Wei, Wang, Schuurmans, Bosma, hsin Chi, Xia, Le, and Zhou]{Wei2022ChainOT}
Jason Wei, Xuezhi Wang, Dale Schuurmans, Maarten Bosma, Ed~Huai hsin Chi, F. Xia, Quoc Le, and Denny Zhou.
\newblock Chain of thought prompting elicits reasoning in large language models.
\newblock \emph{ArXiv}, abs/2201.11903, 2022{\natexlab{b}}.

\bibitem[Williams(2004)]{Williams2004SimpleSG}
Ronald~J. Williams.
\newblock Simple statistical gradient-following algorithms for connectionist reinforcement learning.
\newblock \emph{Machine Learning}, 8:\penalty0 229--256, 2004.

\bibitem[Wu et~al.(2023)Wu, Yin, Qi, Wang, Tang, and Duan]{Wu2023VisualCT}
Chenfei Wu, Sheng-Kai Yin, Weizhen Qi, Xiaodong Wang, Zecheng Tang, and Nan Duan.
\newblock Visual chatgpt: Talking, drawing and editing with visual foundation models.
\newblock \emph{ArXiv}, abs/2303.04671, 2023.

\bibitem[Xu et~al.(2023)Xu, Yang, Lin, Wang, Zhou, Zhang, and Mao]{Xu2023ExpertPromptingIL}
Benfeng Xu, An Yang, Junyang Lin, Quang Wang, Chang Zhou, Yongdong Zhang, and Zhendong Mao.
\newblock Expertprompting: Instructing large language models to be distinguished experts.
\newblock \emph{ArXiv}, abs/2305.14688, 2023.

\bibitem[Yao et~al.(2023{\natexlab{a}})Yao, Yu, Zhao, Shafran, Griffiths, Cao, and Narasimhan]{Yao2023TreeOT}
Shunyu Yao, Dian Yu, Jeffrey Zhao, Izhak Shafran, Thomas~L. Griffiths, Yuan Cao, and Karthik Narasimhan.
\newblock Tree of thoughts: Deliberate problem solving with large language models.
\newblock \emph{ArXiv}, abs/2305.10601, 2023{\natexlab{a}}.

\bibitem[Yao et~al.(2023{\natexlab{b}})Yao, Li, and Zhao]{Yao2023BeyondCE}
Yao Yao, Z. Li, and Hai Zhao.
\newblock Beyond chain-of-thought, effective graph-of-thought reasoning in large language models.
\newblock \emph{ArXiv}, abs/2305.16582, 2023{\natexlab{b}}.

\bibitem[Ye et~al.(2023{\natexlab{a}})Ye, Xu, Xu, Ye, Yan, Zhou, Wang, Hu, Shi, Shi, Li, Xu, Chen, Tian, Qi, Zhang, and Huang]{Ye2023mPLUGOwlME}
Qinghao Ye, Haiyang Xu, Guohai Xu, Jiabo Ye, Ming Yan, Yi Zhou, Junyan Wang, Anwen Hu, Pengcheng Shi, Yaya Shi, Chenliang Li, Yuanhong Xu, Hehong Chen, Junfeng Tian, Qiang Qi, Ji Zhang, and Feiyan Huang.
\newblock mplug-owl: Modularization empowers large language models with multimodality.
\newblock \emph{ArXiv}, abs/2304.14178, 2023{\natexlab{a}}.

\bibitem[Ye et~al.(2023{\natexlab{b}})Ye, Xu, Ye, Yan, Hu, Liu, Qian, Zhang, Huang, and Zhou]{Ye2023mPLUGOwl2RM}
Qinghao Ye, Haiyang Xu, Jiabo Ye, Mingshi Yan, Anwen Hu, Haowei Liu, Qi Qian, Ji Zhang, Fei Huang, and Jingren Zhou.
\newblock mplug-owl2: Revolutionizing multi-modal large language model with modality collaboration.
\newblock \emph{ArXiv}, abs/2311.04257, 2023{\natexlab{b}}.

\bibitem[Zhai et~al.(2023)Zhai, Mustafa, Kolesnikov, and Beyer]{zhai2023sigmoid}
Xiaohua Zhai, Basil Mustafa, Alexander Kolesnikov, and Lucas Beyer.
\newblock Sigmoid loss for language image pre-training.
\newblock In \emph{Proceedings of the IEEE/CVF International Conference on Computer Vision}, pages 11975--11986, 2023.

\bibitem[Zhang et~al.(2023{\natexlab{a}})Zhang, Chen, Bukharin, He, Cheng, Chen, and Zhao]{Zhang2023AdaLoRA}
Qingru Zhang, Minshuo Chen, Alexander~W. Bukharin, Pengcheng He, Yu Cheng, Weizhu Chen, and Tuo Zhao.
\newblock Adaptive budget allocation for parameter-efficient fine-tuning.
\newblock \emph{ArXiv}, abs/2303.10512, 2023{\natexlab{a}}.

\bibitem[Zhang et~al.(2023{\natexlab{b}})Zhang, Han, Zhou, Hu, Yan, Lu, Li, Gao, and Qiao]{Zhang2023LLaMAAdapter1}
Renrui Zhang, Jiaming Han, Aojun Zhou, Xiangfei Hu, Shilin Yan, Pan Lu, Hongsheng Li, Peng Gao, and Yu~Jiao Qiao.
\newblock Llama-adapter: Efficient fine-tuning of language models with zero-init attention.
\newblock \emph{ArXiv}, abs/2303.16199, 2023{\natexlab{b}}.

\bibitem[Zhang et~al.(2022)Zhang, Zhang, Li, and Smola]{Zhang2022AutomaticCO}
Zhuosheng Zhang, Aston Zhang, Mu Li, and Alexander~J. Smola.
\newblock Automatic chain of thought prompting in large language models.
\newblock \emph{ArXiv}, abs/2210.03493, 2022.

\bibitem[Zhang et~al.(2023{\natexlab{c}})Zhang, Zhang, Li, Zhao, Karypis, and Smola]{Zhang2023MultimodalCR}
Zhuosheng Zhang, Aston Zhang, Mu Li, Hai Zhao, George Karypis, and Alexander~J. Smola.
\newblock Multimodal chain-of-thought reasoning in language models.
\newblock \emph{ArXiv}, abs/2302.00923, 2023{\natexlab{c}}.

\bibitem[Zhao et~al.(2023)Zhao, Cai, Si, Ma, An, Chen, Liu, Wang, Han, and Chang]{Zhao2023MMICLEV}
Haozhe Zhao, Zefan Cai, Shuzheng Si, Xiaojian Ma, Kaikai An, Liang Chen, Zixuan Liu, Sheng Wang, Wenjuan Han, and Baobao Chang.
\newblock Mmicl: Empowering vision-language model with multi-modal in-context learning.
\newblock \emph{ArXiv}, abs/2309.07915, 2023.

\bibitem[Zheng et~al.(2023)Zheng, Yang, Tang, Zhou, and Yang]{Zheng2023DDCoTDC}
Ge Zheng, Bin Yang, Jiajin Tang, Hong-Yu Zhou, and Sibei Yang.
\newblock Ddcot: Duty-distinct chain-of-thought prompting for multimodal reasoning in language models.
\newblock \emph{ArXiv}, abs/2310.16436, 2023.

\bibitem[Zhu et~al.(2023)Zhu, Chen, Shen, Li, and Elhoseiny]{zhu2023minigpt4}
Deyao Zhu, Jun Chen, Xiaoqian Shen, Xiang Li, and Mohamed Elhoseiny.
\newblock Minigpt-4: Enhancing vision-language understanding with advanced large language models.
\newblock \emph{arXiv preprint arXiv:2304.10592}, 2023.

\end{thebibliography}
